\PassOptionsToPackage{table}{xcolor} 
\documentclass[runningheads]{llncs}

\usepackage[year=2026,ID=6202]{eccv}

\usepackage{eccvabbrv}

\usepackage{graphicx}
\usepackage{booktabs}
\usepackage{multirow}
\usepackage[accsupp]{axessibility}  

\usepackage[most]{tcolorbox}
\definecolor{graybg}{rgb}{0.95, 0.95, 0.95}
\usepackage{enumitem}

\usepackage{hyperref}

\usepackage{orcidlink}

\begin{document}

\title{From Passive Observer to Active Critic: Reinforcement Learning Elicits Process Reasoning for Robotic Manipulation
} 

\titlerunning{Reinforcement Learning Elicits Process Reasoning for Robotic Manipulation}

\author{
    Yibin Liu\inst{1,2}\thanks{Equal contribution. This work was done during Yibin Liu’s and Yaxing Lyu’s internships at ScaleLab in Shanghai Jiao Tong University.}\orcidlink{0009-0004-2925-7798} \and
    Yaxing Lyu\inst{3}$^\star$\orcidlink{0009-0002-5831-2739} \and
    Daqi Gao\inst{1}$^\star$\orcidlink{0009-0005-7700-2228} \and
    Zhixuan Liang\inst{4}\orcidlink{0009-0008-6815-9866} \and \\
    Weiliang Tang\inst{5}\orcidlink{0009-0000-1220-7031} \and
    Shilong Mu\inst{6}\orcidlink{0009-0004-3638-6539} \and
    Xiaokang Yang\inst{1}\orcidlink{0000-0003-4029-3322} \and
    Yao Mu\inst{1}\thanks{Corresponding author: \email{muyao@sjtu.edu.cn}} \orcidlink{0000-0002-0321-021X}
}

\authorrunning{Y.~Liu et al.}


\institute{
\begin{tabular}{c c }
$^1$ Shanghai Jiao Tong University & $^2$ Northeastern University \\ 
$^3$ Xiamen University Malaysia & $^4$ The University of Hong Kong \\ 
$^5$ The Chinese University of Hong Kong & $^6$ Xspark AI \\
\multicolumn{2}{c}{\email{\{liuyibin@stumail.neu.edu.cn, muyao@sjtu.edu.cn\}}}
\end{tabular}
}

\maketitle

\vspace{-1.0cm}

\setlength{\textfloatsep}{12pt plus 2pt minus 4pt}
\setlength{\floatsep}{10pt plus 2pt minus 2pt}
\setlength{\intextsep}{10pt plus 2pt minus 2pt}

\begin{abstract}

Accurate process supervision remains a critical challenge for long-horizon robotic manipulation. A primary bottleneck is that current video MLLMs, trained primarily under a Supervised Fine-Tuning (SFT) paradigm, function as passive ``Observers'' that recognize ongoing events rather than evaluating the current state relative to the final task goal. In this paper, we introduce \textbf{PRIMO R1} (\textbf{P}rocess \textbf{R}easoning \textbf{I}nduced \textbf{MO}nitoring), a 7B framework that transforms video MLLMs into active ``Critics''. We leverage outcome-based Reinforcement Learning to incentivize explicit Chain-of-Thought generation for progress estimation. Furthermore, our architecture constructs a structured temporal input by explicitly anchoring the video sequence between initial and current state images. Supported by the proposed \textbf{PRIMO Dataset} and \textbf{Benchmark}, extensive experiments across diverse in-domain environments and out-of-domain real-world humanoid scenarios demonstrate that PRIMO R1 achieves state-of-the-art performance. Quantitatively, our 7B model achieves a 50\% reduction in the mean absolute error of specialized reasoning baselines, demonstrating significant relative accuracy improvements over 72B-scale general MLLMs. Furthermore, PRIMO R1 exhibits strong zero-shot generalization on difficult failure detection tasks. We establish state-of-the-art performance on the RoboFail benchmark with 67.0\% accuracy, surpassing closed-source models like OpenAI o1 6.0\%. The project website is: \href{https://10-oasis-01.github.io/primo-r1-website/}{10-oasis-01.github.io/primo-r1-website}.


\end{abstract}

\section{Introduction}

\begin{figure}[t]
    \centering
    \includegraphics[width=1\linewidth]{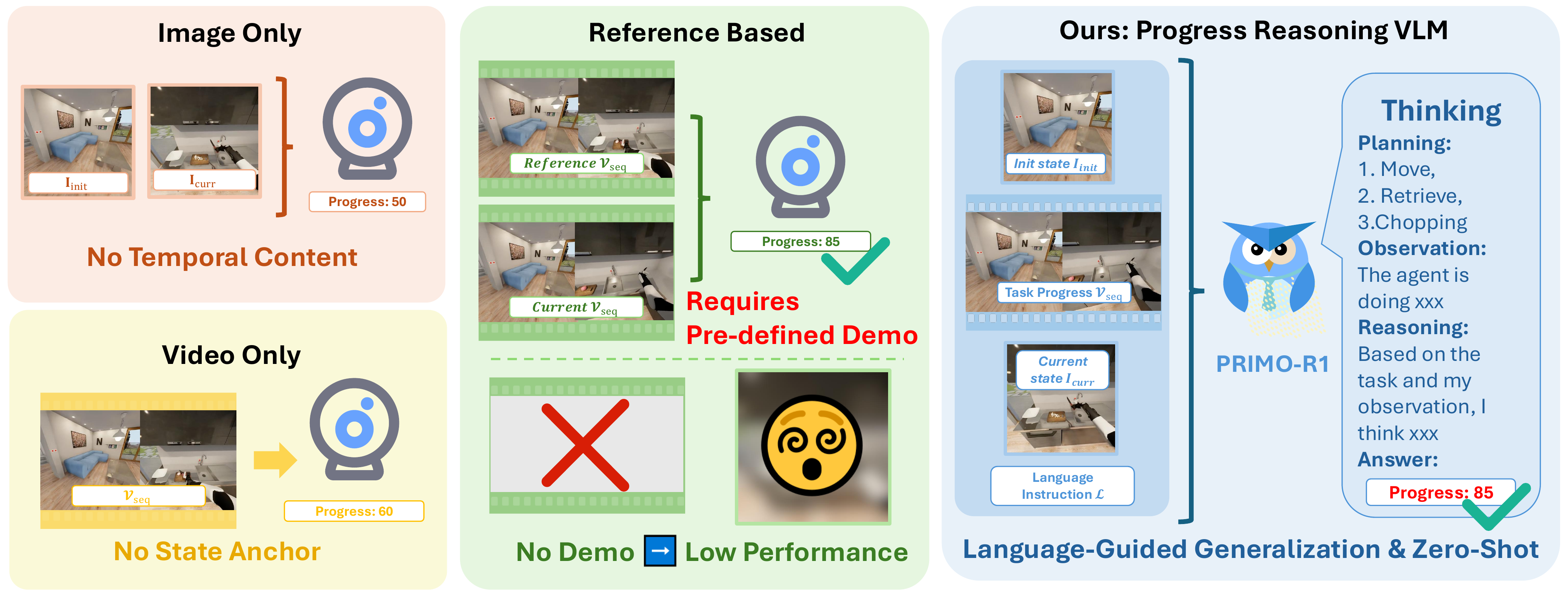}
    \caption{Paradigm comparison: Prior approaches vs. our PRIMO R1.}
    \label{fig:teaser}
    \vspace{-18pt}
\end{figure}

The pursuit of general-purpose robots capable of performing long-horizon manipulation tasks remains a central challenge in embodied AI. A critical bottleneck in acquiring such skills is to derive effective reward signals. While sparse rewards (\textit{e.g.}, binary success/failure) are easy to specify, they are often insufficient for efficient policy learning in complex environments. Conversely, dense rewards, which provide granular feedback on task progress, typically rely on laborious manual engineering or privileged access to ground-truth states unavailable in the real world. Recent advances in Vision-Language Models (VLMs) and Multimodal Large Language Models (MLLMs)~\cite{ma2023liv, ma2022vip, zhai2025vision} have sparked hope for learning universal reward functions directly from visual observations. However, deploying these models as reliable ``process supervisors'' reveals a fundamental limitation in their current paradigm.

Existing video MLLMs, despite their impressive capabilities in captioning and QA, primarily function as passive ``Observers''. They excel at describing \textit{what} is happening but struggle with the rigorous quantitative reasoning required to judge \textit{how well} the task is proceeding. Most prior approaches treat progress estimation as a standard regression or classification problem via supervised fine-tuning. Within this paradigm, models are optimized to recognize and describe ongoing events, rather than to measure the actual distance between the current state and the final task goal. Consequently, these ``Observers'' are brittle: they fail to generalize to unseen objects, cannot explain their predictions, and crucially, often assign high progress scores to failed attempts simply because the visual trajectory resembles a successful motion. This exposes a critical structural deficit: without explicit temporal boundary anchoring and continuous reasoning pathways, models are unable to align continuous visual trajectories with the discrete logical conditions required for task success.

To bridge the gap between passive perception and active evaluation, we argue that a reliable reward model must evolve from an Observer into an active \textbf{``Critic''}. We introduce \textbf{PRIMO R1} (\textbf{P}rocess \textbf{R}easoning \textbf{I}nduced \textbf{MO}nitoring), a 7B model framework that elicits explicit process reasoning from video MLLMs. Instead of supervising the model with a single scalar label, we leverage Reinforcement Learning (RL) to incentivize the generation of Chain-of-Thought. Furthermore, to address the loss of detail in continuous dynamic feature spaces, our architecture employs a structural prompting strategy by explicitly anchoring the video sequence between initial and current state images. This design provides clear visual boundary conditions that transform the reasoning task from generic temporal perception into a structured state-alignment verification. By conditioning this reasoning process on diverse natural language task goals, we establish a structural connection between the objective input and the reasoning execution, effectively exploiting the inherent linguistic generalization capabilities of foundational MLLMs. To support this paradigm, we construct the \textbf{PRIMO Dataset}, encompassing both SFT and RL post-training data with CoT annotations, and \textbf{PRIMO Benchmark}, designed to systematically evaluate out-of-domain generalization across cross-task and cross-environment settings.

Our experiments reveal that optimizing a policy model for continuous progress reasoning intrinsically constructs the temporal context representations required for discrete failure detection. By enforcing rigorous temporal alignment and self-reasoning, PRIMO R1 achieves state-of-the-art performance across multiple domains. Quantitatively, our 7B model attains an average Mean Relative Accuracy (MRA) of 82.90 and a Mean Absolute Error (MAE) of 15.52, effectively outperforming 72B-scale general MLLMs by a margin of +9.10 absolute MRA points. Furthermore, it exhibits robust zero-shot generalization in execution anomaly verification, achieving 67.0\% accuracy on the RoboFail benchmark and surpassing parameter-heavy closed-source models including GPT-4o and OpenAI o1.

Our contributions can be summarized as follows:
\begin{itemize}[topsep=2pt]
    \item We introduce \textbf{PRIMO R1}, a 7B reasoning model that effectively transforms video MLLMs from passive Observers into interpretable Critics. It achieves SOTA performance in task progress estimation and failure detection.
    \item We present the \textbf{PRIMO Dataset} for task progress detection, covering both SFT and RL post-training data with CoT annotations, alongside \textbf{PRIMO Bench}, which systematically evaluates the out-of-domain generalization capabilities of post-training methods in video-based MLLMs.
    \item We propose a structured temporal input strategy that explicitly anchors video sequences between initial and current state images. This boundary anchoring facilitates high-precision state alignment, achieving a 50\% reduction in the mean absolute error compared to specialized baselines.
    \item We demonstrate that optimizing for progress reasoning intrinsically enables robust zero-shot generalization for failure detection. This is validated on the RoboFail benchmark, where PRIMO R1 achieves a state-of-the-art 67.0\% accuracy, surpassing closed-source models like OpenAI o1 by 6.0\%.
\end{itemize}

\section{Related Work}

\subsection{Multimodal Large Language Models for Video Understanding}
Early video MLLMs adapted static architectures via temporal aggregation~\cite{maaz2024video, lin2024video} and mitigated memory bottlenecks through context compression and hierarchical structures~\cite{cheng2024videollama, li2024llama, zhang2024long, ren2025vista, song2024moviechat}. These architectures operate predominantly as passive ``Observers,'' excelling at perceptual QA but lacking quantitative temporal reasoning. The structural transition toward an active ``Critic'' paradigm necessitates explicit temporal localization, prompting recent designs to integrate evidence searching and timestamp encoding~\cite{yu2023self, ren2024timechat, huang2024vtimellm}. Our framework completes this transition by enabling progress reasoning for rigorous temporal judgment.

\subsection{Vision-Based State Estimation and Reward Modeling}
Semantic reward modeling leverages VLMs to encode universal value functions via representation distances~\cite{ma2023liv, ma2022vip} or frozen embeddings~\cite{sontakke2023roboclip}. For explicit progress estimation, recent mechanisms employ frame-ordering~\cite{ma2024vision}, multi-modal integration~\cite{zhai2025vision, zhang2026progresslm}, and synthetic trajectory augmentation~\cite{zhang2025rewind}. A primary structural limitation of VLAC~\cite{zhai2025vision}, Robo-Dopamine~\cite{tan2025robo}, and PROGRESSLM~\cite{zhang2026progresslm} is their functional dependency on explicit reference demonstrations. Furthermore, framing estimation as direct regression via Supervised Fine-Tuning (SFT) restricts the model's capacity for causal failure analysis. Our method bypasses reference dependency by explicitly eliciting process reasoning chains.

\subsection{Reinforcement Learning for Reasoning Elicitation}
Inference-time scaling and outcome-based Reinforcement Learning (RL) induce Chain-of-Thought (CoT) behaviors without dense annotations~\cite{guo2025deepseek}. This ``R1 paradigm'' has expanded into multimodal domains, enhancing static visual reasoning~\cite{shen2025vlm, yang2025r1} and dynamic temporal grounding~\cite{feng2025video, wang2025time}. Parallel architectural optimizations include dynamic frame sampling~\cite{ge2025framemind} and current-state image anchoring for planning~\cite{chen2025grpo}. We map this capability to robotic process supervision, formulating task completion metrics as outcome rewards to elicit verifiable, self-correcting reasoning paths for task assessment.

\section{Method}

\begin{figure}[t]
    \centering
    \includegraphics[width=1\linewidth]{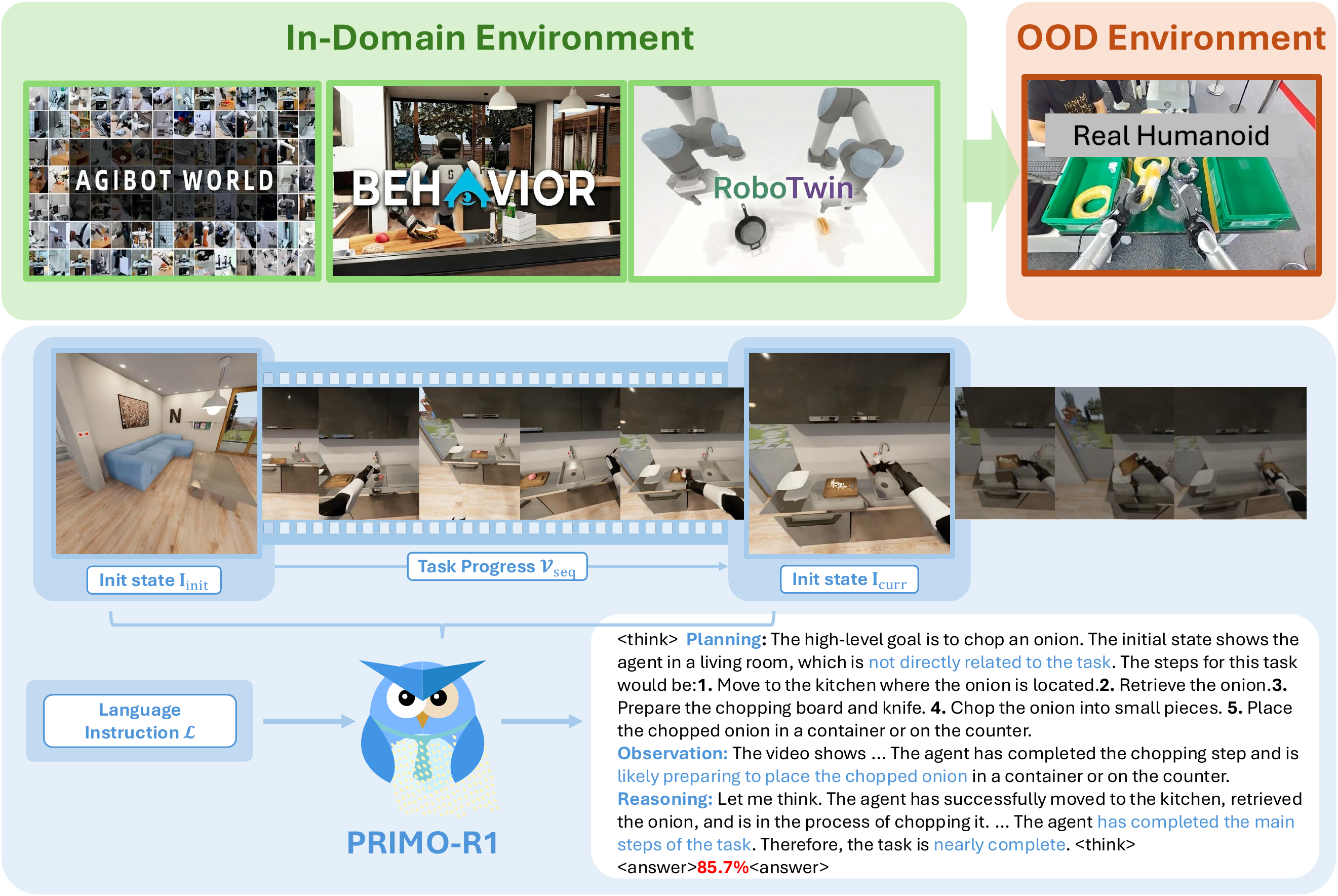}
    \caption{\textbf{Overall framework of PRIMO R1.} Evaluated across in-domain simulations (AgiBot, BEHAVIOR, RoboTwin) and OOD real humanoid environments, the model processes a video sequence ($V_{seq}$) anchored by initial ($I_{init}$) and current ($I_{curr}$) states. It generates an explicit Chain-of-Thought to output the final progress estimate.}
    \label{fig:Overall}
    \vspace{-18pt}
\end{figure}

\subsection{Problem Formulation}

We formalize the task of robotic process supervision as a multi-modal state estimation problem. The input space consists of four key variables: an initial state image $I_{init}$ capturing the environment prior to execution, a process video sequence $V_{seq} = \{v_1, v_2, \dots, v_T\}$ representing the temporal state transitions, a current state image $I_{curr}$ reflecting the latest observed outcome, and a language instruction $\mathcal{I}$ specifying the task goal (the specific structure and content of $\mathcal{I}$ are detailed in Figure~\ref{fig:prompt_template}). The objective is to learn a mapping function $F$ that evaluates the visual tuple $(I_{init}, V_{seq}, I_{curr})$ conditioned on the instruction $\mathcal{I}$ as the semantic reference, outputting a scalar progress indicator $y \in [0, 100]$, where $y=0$ denotes the initial state and $y=100$ signifies successful state. As demonstrated in our ablation study (Table~\ref{tab:ablation_temporal}), explicitly modeling both boundary states ($I_{init}$ and $I_{curr}$) alongside the temporal transition ($V_{seq}$) is a necessary structural prerequisite for accurate progress estimation across varying task horizons.

In the standard paradigm, existing video MLLMs function as passive ``Observers'', treating progress estimation as a direct regression or classification task. They directly model the distribution of the target $y_{gt}$ conditioned on the input tuple $(I_{init}, V_{seq}, I_{curr}, \mathcal{I})$ via Supervised Fine-Tuning (SFT). This formulation isolates visual features at a surface level, bypassing the underlying causal structure of the state transitions.

To transform the model into an active ``Critic'', we reformulate the prediction process from direct scalar regression into a multi-step generative reasoning task. We define a policy $\pi_\theta$ that sequentially generates a latent reasoning chain (Chain-of-Thought) $\mathcal{C}$, followed by the final progress estimate $\hat{y}$. Rather than relying on dense annotations to supervise the intermediate variable $\mathcal{C}$, we optimize $\pi_\theta$ using Reinforcement Learning. The optimization objective maximizes the expected reward $R(\hat{y}, y_{gt})$, which is computed solely based on the accuracy of the final prediction $\hat{y}$. This structural dependency incentivizes the policy to self-organize the intermediate reasoning $\mathcal{C}$ to accurately align the temporal transition $V_{seq}$ between the boundary states $I_{init}$ and $I_{curr}$. Crucially, conditioning this generative reasoning process on diverse natural language task goals ($\mathcal{I}$) establishes a direct structural mapping between the semantic objective and the visual execution logic, explicitly exploiting the linguistic generalization capabilities of foundational MLLMs to process varying evaluation criteria. The complete architectural workflow of this framework is illustrated in Fig.~\ref{fig:Overall}.

\begin{figure}[t]
    \centering
    \includegraphics[width=1\linewidth]{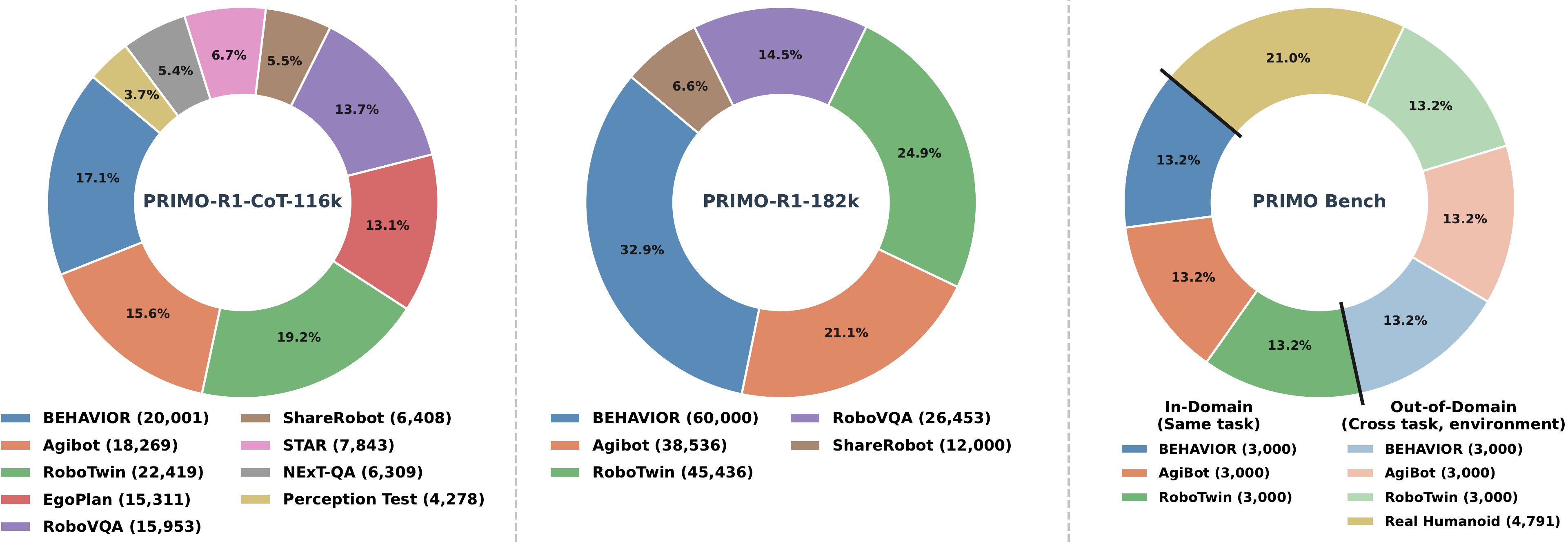}
    \caption{\textbf{Dataset distribution for SFT (left), RL (middle), and PRIMO Bench (right).} Charts show sample counts and domain percentages (e.g., BEHAVIOR, AgiBot, RoboTwin). The PRIMO Bench highlights the data split between In-Domain and Out-of-Domain evaluation sets. See Appendix~\ref{sec:appendix_datasets} for details.}
    \label{fig:dataset_dist}
\end{figure}

\subsection{PRIMO Dataset and Benchmark}
\label{sec:dataset_and_bench}

To systematically elicit and evaluate the temporal reasoning capabilities of Video MLLMs for robotic process supervision, we present the \textbf{PRIMO Dataset} and the accompanying \textbf{PRIMO Bench}.

\vspace{1mm}
\noindent\textbf{PRIMO Dataset for Post-Training.} 
The PRIMO Dataset is meticulously constructed to support our two-stage post-training paradigm, covering both Supervised Fine-Tuning (SFT) and Reinforcement Learning (RL) data. Unlike standard video QA datasets, our data features fine-grained progress indicators annotated with Chain-of-Thought reasoning paths. The training corpus aggregates multi-source trajectories from a real-world environment (AgiBot) and two high-fidelity simulations (BEHAVIOR-1k and RoboTwin). Additionally, to maintain data diversity during the SFT phase, we incorporate several general video reasoning datasets to augment the training corpus, yielding a comprehensive collection partitioned into a 116k-sample SFT dataset (PRIMO-R1-CoT-116k) and a 182k-sample RL dataset (PRIMO-R1-182k), as illustrated in Figure~\ref{fig:dataset_dist}.

\vspace{1mm}
\noindent\textbf{PRIMO Bench for Generalization Evaluation.}
To systematically evaluate the robustness of post-training methods against varying degrees of distribution shift, we introduce \textbf{PRIMO Bench}, which categorizes evaluation into two splits:
\begin{itemize}
    \item \textbf{In-Domain (ID) - Same Task:} Evaluates the model's estimation accuracy on task categories that were exposed during the training phase within the three seen environments.
    \item \textbf{Out-of-Domain (OOD) - Cross-Task \& Cross-Environment:} Designed to test zero-shot generalization. \textit{Cross-Task} evaluates the model on entirely unseen tasks within the familiar environments. \textit{Cross-Environment} introduces a stringent unseen domain transfer challenge, evaluating the model on real-world trajectories collected via teleoperation of a different physical humanoid robot (Leju KUAVO-MY) in unstructured physical environments (e.g., factories and service scenarios).
\end{itemize}

Comprehensive details regarding raw data collection, semantic annotation synthesis, data processing methodologies, and exact dataset statistics of our dataset and benchmark are provided in Appendix~\ref{sec:appendix_datasets}.

\subsection{Process Reasoning RL with Group Relative Policy Optimization}
\label{sec:grpo}

To transform the Video MLLM from a passive observer into an active critic capable of self-correction, we employ Group Relative Policy Optimization (GRPO)~\cite{guo2025deepseek}. Unlike standard Proximal Policy Optimization (PPO)~\cite{schulman2017proximal}, which relies on a computationally expensive value function critic to estimate the baseline, GRPO leverages the group statistics of sampled outputs to estimate the baseline. This is particularly advantageous for Video MLLMs, where the memory overhead of maintaining a separate value network alongside the policy model is prohibitive.

\vspace{1mm}
\noindent\textbf{Group Sampling and Advantage Estimation.}
Formally, for a given task tuple $(I_{init}, V_{seq}, I_{curr}, \mathcal{I})$, we sample a group of $G$ outputs $\{o_1, o_2, \dots, o_G\}$ from the policy $\pi_{\theta_{old}}$. Each output $o_i$ consists of a reasoning chain $\mathcal{C}_i$ (enclosed in \texttt{<think>} tags) and a final progress estimate $\hat{y}_i$.
Instead of training a value function $V(x)$, GRPO computes the advantage $A_i$ for each output $o_i$ by normalizing its reward $r_i$ against the group's distribution:
{\vspace{-3pt}\small
\begin{equation}
    A_i = \frac{r_i - \text{mean}(\{r_1, \dots, r_G\})}{\text{std}(\{r_1, \dots, r_G\}) + \epsilon},
\end{equation}
}
where $\epsilon$ is a small constant for numerical stability. This relative advantage encourages the model to generate reasoning paths that yield higher rewards than the average of its current stochastic explorations, effectively filtering out "hallucinated" progress estimates.

\vspace{1mm}
\noindent\textbf{Rule-Based Reward Design.}
A core challenge in eliciting reasoning is defining an effective reward signal without dense annotation. We define a composite reward function $R(o_i, y_{gt}) = r_{\text{fmt}} + r_{\text{acc}}$ targeting both structure and precision:

\noindent\textbf{(1) Format Reward ($r_{\text{fmt}}$).} 
To explicitly induce a Chain-of-Thought, we enforce a strict structural constraint. The model receives a positive reward (e.g., $+1$) only if the output strictly follows the pattern \texttt{\textless think\textgreater reasoning\textless/think\textgreater} followed by \texttt{\textless answer\textgreater prediction\textless/answer\textgreater}. This prevents the policy from collapsing into direct guessing.

\noindent\textbf{(2) Accuracy Reward ($r_{\text{acc}}$).} 
Since the task progress $y$ is continuous, treating it as a binary outcome is insufficient. 
To provide dense feedback for numerical reasoning, we adopt a \textit{bounded linear decay} reward function:
{\vspace{-3pt}\small
\begin{equation}
r_{\text{acc}} = \max \left( 0, 1 - \frac{|\hat{y}_i - y_{gt}|}{R_{\text{max}}} \right),
\end{equation}}
where $R_{\text{max}}$ (e.g., $100.0$) represents the maximum error range. 
This formulation ensures the reward starts at $1.0$ for an exact match and linearly decreases to $0.0$ as the error approaches $R_{\text{max}}$, strictly confining the score to the $[0, 1]$ interval.

\vspace{1mm}
\noindent\textbf{Optimization Objective.}
The policy $\pi_\theta$ is updated to maximize the expected advantage while remaining close to the reference policy $\pi_{\text{ref}}$ to prevent reward hacking or language degeneration. The GRPO objective is formulated as:
{\vspace{-5pt}\small
\begin{equation}
\begin{aligned}
    \mathcal{L}_{\text{GRPO}}(\theta) = -\frac{1}{G} \sum_{i=1}^G & \left[ \min \left( \rho_i A_i, \text{clip}(\rho_i, 1-\epsilon, 1+\epsilon) A_i \right) \right. \\
    & \left. - \beta \cdot \mathbb{D}_{\text{KL}} \left( \pi_\theta(o_i|x) || \pi_{\text{ref}}(o_i|x) \right) \right],
\end{aligned}
\end{equation}
}
where $\rho_i = \frac{\pi_\theta(o_i|x)}{\pi_{\theta_{old}}(o_i|x)}$ is the probability ratio, and $\beta$ controls the strength of the KL divergence penalty. By optimizing this objective, PRIMO R1 implicitly learns that generating detailed, causal reasoning in $\mathcal{C}$ is the most reliable strategy to maximize the accuracy reward in $\hat{y}$, thereby emerging as a robust Critic.


\section{Experiments}

In this section, we systematically evaluate the performance of PRIMO R1. The experiments are structured to assess two primary capabilities: continuous task progress estimation across both in-domain simulations and out-of-domain real-world environments, and zero-shot generalization in discrete execution failure detection. Furthermore, we conduct ablation studies to isolate the impact of temporal context modalities on estimation accuracy, followed by qualitative case studies analyzing the structural logic of the generated reasoning chains. Throughout our evaluations, we employ Qwen2.5-VL-7B-Instruct as the foundation model for all training phases. Detailed experimental setups, including hardware specifications, training parameters, and inference configurations like sampled frame count and frame resolution, are provided in Appendix~\ref{sec:appendix_setup}.

\subsection{Evaluation Metrics}
\label{sec:metrics}

We evaluate task progress estimation using Mean Relative Accuracy (MRA) and Mean Absolute Error (MAE).

\textbf{Mean Relative Accuracy (MRA).}
Given a prediction $\hat{y}$, ground-truth progress $y$, and a set of accuracy thresholds $\mathcal{T}$, Mean Relative Accuracy (MRA) is defined as
{\small
\begin{equation}
\mathrm{MRA} =
\frac{1}{|\mathcal{T}|}
\sum_{\tau \in \mathcal{T}}
\mathbb{I}\left(\frac{|\hat{y} - y|}{|y|} < 1 - \tau\right),
\end{equation}}
where $\mathbb{I}(\cdot)$ denotes the indicator function.

\textbf{Mean Absolute Error (MAE).}
MAE is defined as
{\small
\begin{equation}
\mathrm{MAE} = \mathbb{E}\big[|\hat{y} - y|\big],
\end{equation}
}
and is reported to provide a clear measure of absolute prediction error.

\subsection{Main Results: Generalization in Progress Estimation}

We present the evaluation of our proposed method against state-of-the-art baselines. The results are analyzed in two parts: a comprehensive performance comparison across all domains (Table~\ref{tab:main_results}) and an ablation study focusing on the impact of SFT and RL strategies on generalization (Table~\ref{tab:ablation_generalization}).

\begin{table}[t]
\centering
\caption{\textbf{Comparison on Progress Estimation.} We report the Mean Relative Accuracy (MRA $\uparrow$, higher is better) and Mean Absolute Error (MAE $\downarrow$, lower is better) across four distinct environments. The best results are highlighted in \textbf{bold}. Specialized progress and reward models are evaluated under their own recommended input configurations and are therefore not strictly comparable cell-for-cell with models run under our unified protocol; the ProgressLM-3B-RL row in the Reasoning \& Video group is the same model re-evaluated under our unified setting.}
\label{tab:main_results}
\resizebox{\textwidth}{!}{%
\begin{tabular}{l|cc|cc|cc|cc|cc}
\toprule
\multirow{2}{*}{\textbf{Model}} & \multicolumn{2}{c|}{\textbf{AgiBot}} & \multicolumn{2}{c|}{\textbf{Behavior}} & \multicolumn{2}{c|}{\textbf{RoboTwin}} & \multicolumn{2}{c|}{\textbf{Real Humanoid}} & \multicolumn{2}{c}{\textbf{Average}} \\
 & MRA ($\uparrow$) & MAE ($\downarrow$) & MRA ($\uparrow$) & MAE ($\downarrow$) & MRA ($\uparrow$) & MAE ($\downarrow$) & MRA ($\uparrow$) & MAE ($\downarrow$) & MRA ($\uparrow$) & MAE ($\downarrow$) \\
\midrule
\multicolumn{11}{l}{\textit{Closed-Source Models}} \\
GPT-5 mini & 74.27 & 24.81 & 79.60 & 20.08 & 80.52 & 18.34 & 67.14 & 32.59 & 75.38 & 23.96 \\
GPT-4o & 81.01 & 18.99 & 79.92 & 20.08 & 81.73 & 18.27 & 74.65 & 25.35 & 79.33 & 20.67 \\
Gemini 2.5 Flash & 73.54 & 26.41 & 78.01 & 21.99 & 81.04 & 18.58 & 67.37 & 32.63 & 74.99 & 24.90 \\
Claude-Haiku-4.5 & 74.40 & 25.59 & 70.93 & 29.07 & 74.13 & 25.87 & 72.68 & 27.32 & 73.04 & 26.96 \\
\midrule
\multicolumn{11}{l}{\textit{Open-Source General MLLMs}} \\
Qwen2.5-VL-7B & 77.43 & 22.56 & 69.91 & 30.06 & 67.37 & 32.62 & 56.46 & 34.73 & 67.79 & 29.99 \\
InternVL 3.5 8B & 78.52 & 21.47 & 72.19 & 27.15 & 70.81 & 29.18 & 65.44 & 34.55 & 71.74 & 28.09 \\
Qwen2.5-VL-72B & 78.00 & 22.00 & 79.49 & 20.50 & 75.41 & 24.59 & 62.29 & 28.10 & 73.80 & 23.80 \\
\midrule
\multicolumn{11}{l}{\textit{Reasoning \& Video MLLMs}} \\
ProgressLM-3B-RL & 42.90 & 30.83 & 46.43 & 28.23 & 31.84 & 36.48 & 32.00 & 34.90 & 38.29 & 32.61 \\
Video R1 7B & 72.42 & 27.58 & 70.63 & 29.37 & 70.86 & 29.14 & 53.57 & 31.87 & 66.87 & 29.49 \\
Robobrain 7B & 72.99 & 25.91 & 72.52 & 26.97 & 70.41 & 28.85 & 55.83 & 28.51 & 67.94 & 27.56 \\
Cosmos-Reasoning 7B & 72.48 & 27.01 & 67.06 & 32.35 & 73.14 & 25.85 & 59.39 & 31.41 & 66.52 & 29.12 \\
\midrule
\multicolumn{11}{l}{\textit{Specialized Progress \& Reward Models (own input configuration)}} \\
VLAC & 84.92 & 15.08 & 76.86 & 23.14 & 67.04 & 32.96 & 70.77 & 29.23 & 74.90 & 25.10 \\
GVL (Gemini 2.5 Flash) & 56.81 & 35.84 & 58.11 & 32.31 & 61.18 & 27.91 & 48.77 & 38.37 & 56.22 & 33.61 \\
Robo-Dopamine-3B & 81.18 & 18.82 & 57.78 & 42.22 & 82.96 & 17.04 & 68.43 & 31.57 & 72.59 & 27.41 \\
Robo-Dopamine-4B & 83.88 & 16.12 & 65.30 & 34.70 & 67.22 & 32.78 & 73.89 & 26.11 & 72.57 & 27.43 \\
ProgressLM & 81.13 & 18.79 & 80.89 & 15.97 & 67.03 & 32.97 & \textbf{84.24} & \textbf{15.76} & 78.32 & 20.87 \\
\midrule
\rowcolor{graybg} \textbf{PRIMO R1 (Ours)} & \textbf{87.67} & \textbf{12.33} & \textbf{87.08} & \textbf{12.90} & \textbf{84.52} & \textbf{15.48} & 72.32 & 21.37 & \textbf{82.90} & \textbf{15.52} \\
\bottomrule
\end{tabular}%
}
\end{table}

\vspace{1mm}
\noindent\textbf{Overall Performance.}
Table~\ref{tab:main_results} summarizes the performance of all models across the four evaluation environments: AgiBot, Behavior, RoboTwin, and Real Humanoid. Using the metrics defined in Sec.~\ref{sec:metrics}, PRIMO R1 achieves the highest average MRA (82.90) and the lowest average MAE (15.52), consistently outperforming all evaluated open-source baselines.

Compared with general open-source MLLMs, PRIMO demonstrates a clear advantage. Despite using only a 7B backbone, it surpasses Qwen2.5-VL-72B by 9.10 MRA points (82.90 vs. 73.80). Against specialized reasoning and video MLLMs such as Video R1 7B and Robobrain 7B, PRIMO reduces the average MAE from approximately 27--29 to 15.52, nearly halving the estimation error.

We further compare PRIMO R1 with specialized progress estimation models, including VLAC, GVL, Robo-Dopamine, and ProgressLM, each evaluated using its recommended input configuration. PRIMO R1 still achieves the best overall performance, improving the average MRA from 78.32 to 82.90 while reducing the average MAE from 20.87 to 15.52. It consistently leads on AgiBot, Behavior, and RoboTwin across both metrics. The only exception is the Real Humanoid setting, where ProgressLM achieves higher MRA (84.24 vs. 72.32) and lower MAE (15.76 vs. 21.37), likely benefiting from its dedicated real-world input configuration. In contrast, PRIMO R1 adopts a unified input protocol across all four environments yet still achieves the best average performance, suggesting that explicit process reasoning generalizes more reliably across heterogeneous domains than configuration-specific designs.

\begin{figure}[t]
    \centering
    \includegraphics[width=1\textwidth]{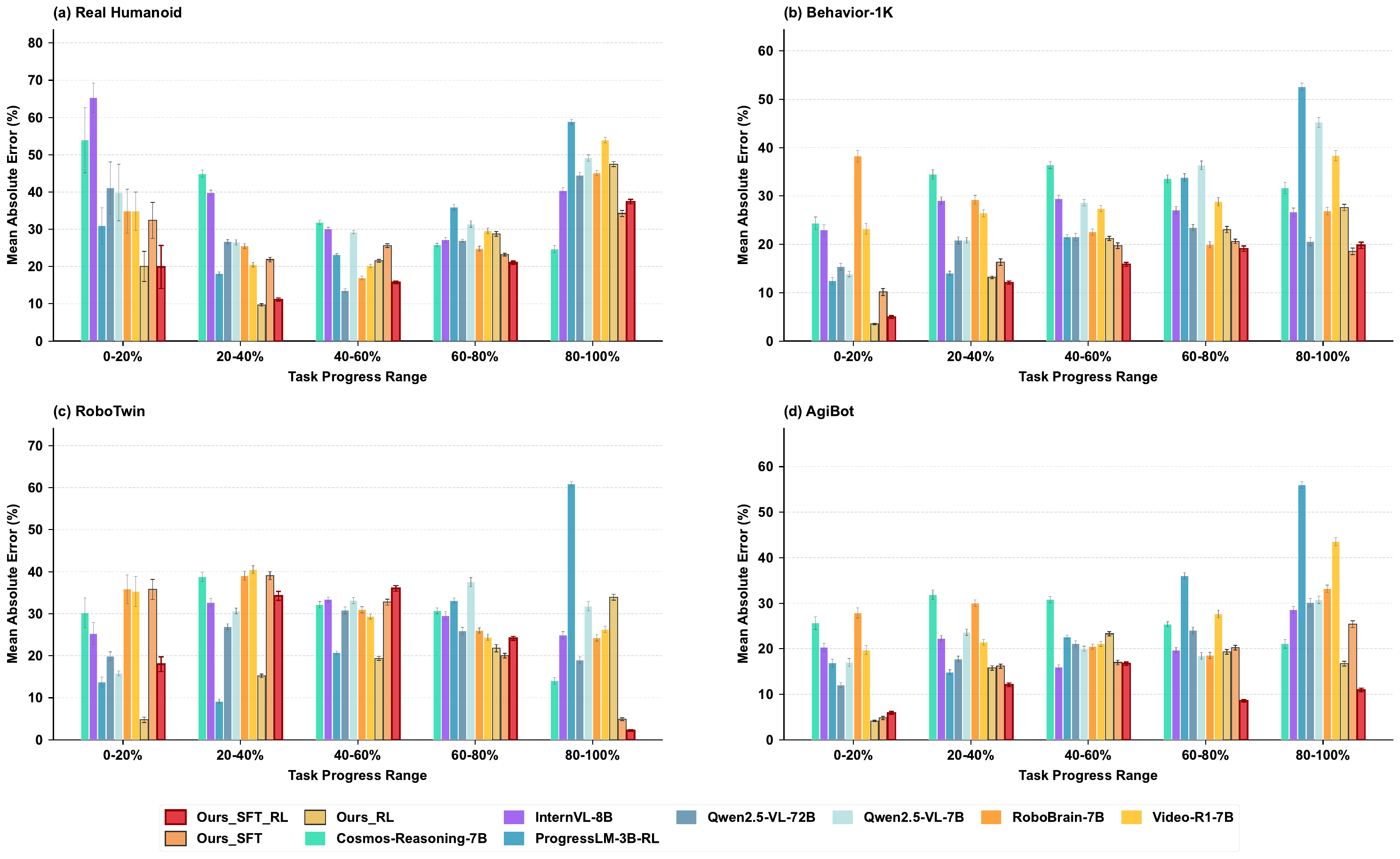}
    \caption{\textbf{Fine-Grained Error Analysis Across Task Progress Intervals.} MAE evaluation across five completion stages in four environments ((a)-(d)). Compared to baselines, our RL-finetuned model (Ours\_SFT\_RL) maintains lower error rates, particularly mitigating severe hallucinations in the final execution stage ($80{-}100\%$).}
    \label{fig:mae_by_progress}
\end{figure}

Figure~\ref{fig:mae_by_progress} further analyzes MAE across five task completion intervals. While baseline models exhibit pronounced error spikes during the final execution stage (\$80{-}100\%\$), PRIMO maintains consistently low errors throughout the trajectory. This indicates that explicit process reasoning effectively mitigates premature completion hallucinations caused by superficial visual similarities near the end of a task. Notably, in the unseen and highly unstructured Real Humanoid environment, PRIMO achieves an MRA of 72.32, substantially outperforming general MLLMs (e.g., Qwen2.5-VL-7B: 56.46). This result highlights the effectiveness of generating an explicit reasoning chain before prediction for improving Sim-to-Real generalization.

\vspace{1mm}
\noindent\textbf{Impact of RL on Generalization.}
To validate the effectiveness of our training pipeline, we analyze the performance evolution from the Base model to the SFT stage, and finally to the RL-finetuned stage. Table~\ref{tab:ablation_generalization} details the performance on In-Domain (ID) seen tasks versus Out-of-Distribution (OOD) unseen tasks.

\begin{table}[t]
\centering
\caption{\textbf{Ablation and Generalization Analysis.} All reported metrics represent Mean Relative Accuracy (MRA $\uparrow$). We compare the Base model, SFT-only model, RL-only model, and our final model (SFT + RL). The results are split into In-Domain (ID) tasks and Out-of-Domain (OOD) tasks to highlight generalization capabilities.}
\label{tab:ablation_generalization}
\resizebox{\textwidth}{!}{%
\begin{tabular}{l|ccc|ccc|c|c}
\toprule
\multirow{3}{*}{\textbf{Model}} & \multicolumn{3}{c|}{\textbf{In-Domain (ID)}} & \multicolumn{4}{c|}{\textbf{Out-of-Domain (OOD)}} & \multirow{3}{*}{\textbf{Avg.}} \\
\cmidrule(lr){2-4} \cmidrule(lr){5-8}
 & \multicolumn{3}{c|}{\textit{Seen Tasks}} & \multicolumn{3}{c|}{\textit{Cross-Task}} & \multicolumn{1}{c|}{\textit{Cross-Environment}} & \\
 & Agibot & Behavior & RoboTwin & Agibot & Behavior & RoboTwin & Real Humanoid & \\
\midrule
Qwen2.5-VL-7B (Base) & 70.83 & 69.13 & 71.19 & 74.45 & 77.47 & 61.01 & 48.12 & 67.46 \\
Our Model (SFT) & 83.37 & 80.38 & 80.63 & 82.02 & 82.61 & 79.13 & 67.30 & 79.35 \\
Our Model (RL) & 86.05 & 85.82 & 73.27 & 82.95 & 81.39 & 75.43 & 52.12 & 76.72 \\
\rowcolor{graybg} \textbf{PRIMO R1 (SFT+RL)} & \textbf{87.83} & \textbf{89.42} & \textbf{88.15} & \textbf{87.67} & \textbf{87.08} & \textbf{84.52} & \textbf{72.32} & \textbf{85.28} \\
\bottomrule
\end{tabular}%
}
\end{table}

The Base Qwen2.5-VL-7B model exhibits a weak zero-shot capability for precise progress estimation (average MRA of $67.46$). While Supervised Fine-Tuning (SFT) significantly improves overall performance to $79.35$, it primarily overfits to the semantic features of the training distribution. This behavior is consistent with the observation that SFT tends to memorize training formats rather than acquire transferable reasoning, whereas RL generalizes to out-of-distribution inputs~\cite{chu2025sft}. The effect is evident in the Cross-Environment (Real Humanoid) OOD setting, where the SFT model only achieves $67.30$.

Interestingly, applying RL directly without SFT (RL-only) yields suboptimal results ($76.72$ average MRA), as the model struggles to autonomously discover the correct output format and structural reasoning paths from scratch. However, the integration of Group Relative Policy Optimization (GRPO) after the SFT phase enables our complete PRIMO (SFT+RL) pipeline to create a powerful synergy. The RL phase pushes ID performance to near $90\%$ (e.g., $89.42\%$ on Behavior) and, more importantly, drastically enhances generalization capabilities. The self-correction and rigorous causal reasoning learned via RL transfer effectively to OOD settings, boosting Cross-Task performance across all simulated environments and lifting the Cross-Environment accuracy to $72.32\%$. This confirms that RL process supervision fundamentally shifts the model from a passive pattern-matcher to an active, generalizing critic.


\begin{table}[t]
\centering
\caption{\textbf{Failure Detection Capabilities.} Accuracy (\%) on the RoboFail benchmark\cite{liu2023reflect}. The evaluation measures the capability of models to effectively detect and quantify task execution failures. Benchmark details are provided in Appendix~\ref{sec:appendix_datasets}.}
\label{tab:robofail_1x3_resized}
\renewcommand{\arraystretch}{1.1}
\resizebox{\textwidth}{!}{%
\begin{tabular}{lc|lc|lc}
\toprule
\textbf{Model} & \textbf{RoboFail} ($\uparrow$) & \textbf{Model} & \textbf{RoboFail} ($\uparrow$) & \textbf{Model} & \textbf{RoboFail} ($\uparrow$) \\
\midrule
\multicolumn{2}{l|}{\textit{Closed-Source}} & \multicolumn{2}{l|}{\textit{Open-Source}} & \multicolumn{2}{l}{\textit{Ours}} \\
Gemini 2.0 Flash & \textbf{67.0} & Qwen2.5-VL-7B      & 57.6 & PRIMO (SFT) & 51.0 \\
GPT-4o           & 63.0 & Nemotron-H-56B     & 64.0 & PRIMO (RL)  & 63.0 \\
OpenAI o1        & 61.0 & Cosmos-Reason1-7B  & 60.0 & \cellcolor{graybg}\textbf{PRIMO R1} & \cellcolor{graybg}\textbf{67.0} \\
Claude-haiku-4.5 & 59.0  & Cosmos-Reason1-56B & 66.2 &             &     \\
\bottomrule
\end{tabular}%
}
\vspace{-12pt}
\end{table}

\subsection{Generalization Enhancement in Failure Detection}

For a Vision-Language Model engaged in process supervision, tracking continuous task progress and detecting discrete execution failures represent coupled dimensions of temporal reasoning. The capability to identify physical constraints or execution errors structurally depends on an underlying representation of intended state transitions. To evaluate the zero-shot generalization of this capability, we test our model on the RoboFail benchmark (details in Appendix~\ref{sec:appendix_datasets}), a completely unseen dataset designed to evaluate "action affordance" and "task completion verification" under complex physical constraints.

Table~\ref{tab:robofail_1x3_resized} details the quantitative performance across different model architectures. The base Qwen2.5-VL-7B model exhibits a baseline accuracy of 57.6\%. Applying Supervised Fine-Tuning (SFT) alone results in a performance regression to 51.0\%. We attribute this drop to format overfitting rather than a limitation of the data: because next-token prediction weights all tokens equally, SFT tends to pattern-match the chain-of-thought structure of the training distribution instead of reasoning over unseen inputs, and prior studies report the same effect, where SFT memorizes formats while RL generalizes~\cite{chu2025sft, feng2025video}. The integration of Process Supervision RL with GRPO corrects this degradation, elevating the accuracy to 63.0\%. By optimizing against outcome-based rewards rather than fixed token sequences, RL recovers genuine reasoning and transfers it to this unseen benchmark. The final PRIMO R1 formulation achieves an accuracy of 67.0\%, matching the closed-source Gemini 2.0 Flash and outperforming larger parameter models, including GPT-4o (63.0\%), OpenAI o1 (61.0\%), and Cosmos-Reason1-56B (66.2\%).

Prior benchmark analyses, such as those conducted for Cosmos-Reason1, indicate that standard Reinforcement Learning targeting physical AI yields limited improvements on RoboFail. The core difficulty of the benchmark stems from the prerequisite for highly observant perception and comprehensive temporal context processing, which operate as distinct variables alongside static physical common sense. The performance delta between Cosmos-Reason1-7B (60.0\%) and PRIMO R1 (67.0\%) establishes a specific functional relationship: optimizing a policy model for continuous progress reasoning explicitly constructs the temporal context representations necessary for failure verification. The capacity for embodied error correction structurally necessitates process reasoning capability as a parallel condition to physical common sense.

\begin{table}[t]
\centering
\caption{\textbf{Ablation on Input Modalities.} We analyze the necessity of temporal context by varying the input information. 
\textbf{$I_{init}$}: Initial state image. 
\textbf{$V_{seq}$}: Process video clip. 
\textbf{$I_{curr}$}: Current state image. 
Results show that temporal context ($V_{seq}$) is crucial for reducing error (MAE).}
\label{tab:ablation_temporal}
\resizebox{1\textwidth}{!}{%
\begin{tabular}{ccc|cc|cc|cc|cc}
\toprule
\multicolumn{3}{c|}{\textbf{Input Modality}} 
& \multicolumn{2}{c|}{\textbf{Agibot}} 
& \multicolumn{2}{c|}{\textbf{Behavior}} 
& \multicolumn{2}{c|}{\textbf{Robotwin}} 
& \multicolumn{2}{c}{\textbf{Avg}} \\

$I_{init}$ & $V_{seq}$ & $I_{curr}$ 
& MAE $\downarrow$ & Acc@10 $\uparrow$ 
& MAE $\downarrow$ & Acc@10 $\uparrow$
& MAE $\downarrow$ & Acc@10 $\uparrow$
& MAE $\downarrow$ & Acc@10 $\uparrow$ \\
\midrule

 &  & \checkmark
 & 59.64 & 0.00
 & 51.91 & 8.17
 & 66.94 & 0.77
 & 59.50 & 2.98 \\

\checkmark &  & \checkmark
 & 43.97 & 18.52
 & 49.59 & 9.73
 & 45.93 & 11.45
 & 46.50 & 13.23 \\

\midrule

 & \checkmark & 
 & 27.58 & 31.34
 & 34.85 & 18.33
 & 47.41 & 8.07
 & 36.61 & 19.25 \\

 & \checkmark & \checkmark
 & 25.04 & \textbf{35.29}
 & 27.59 & 29.21
 & \textbf{40.24} & 17.37
 & 30.96 & 27.29 \\

\checkmark & \checkmark & 
 & \textbf{24.94} & 33.98
 & 32.55 & 23.01
 & 45.37 & 11.97
 & \textbf{34.29} & 22.99 \\

\midrule

\rowcolor{graybg}
\checkmark & \checkmark & \checkmark
 & 29.39 & 27.15
 & \textbf{22.73} & \textbf{31.83}
 & 42.16 & \textbf{27.69}
 & 31.43 & \textbf{28.89} \\

\bottomrule
\end{tabular}
}
\end{table}

\subsection{Ablation Study: The Necessity of Temporal Context}

\begin{figure}[t]
    \centering
    \includegraphics[width=1\linewidth]{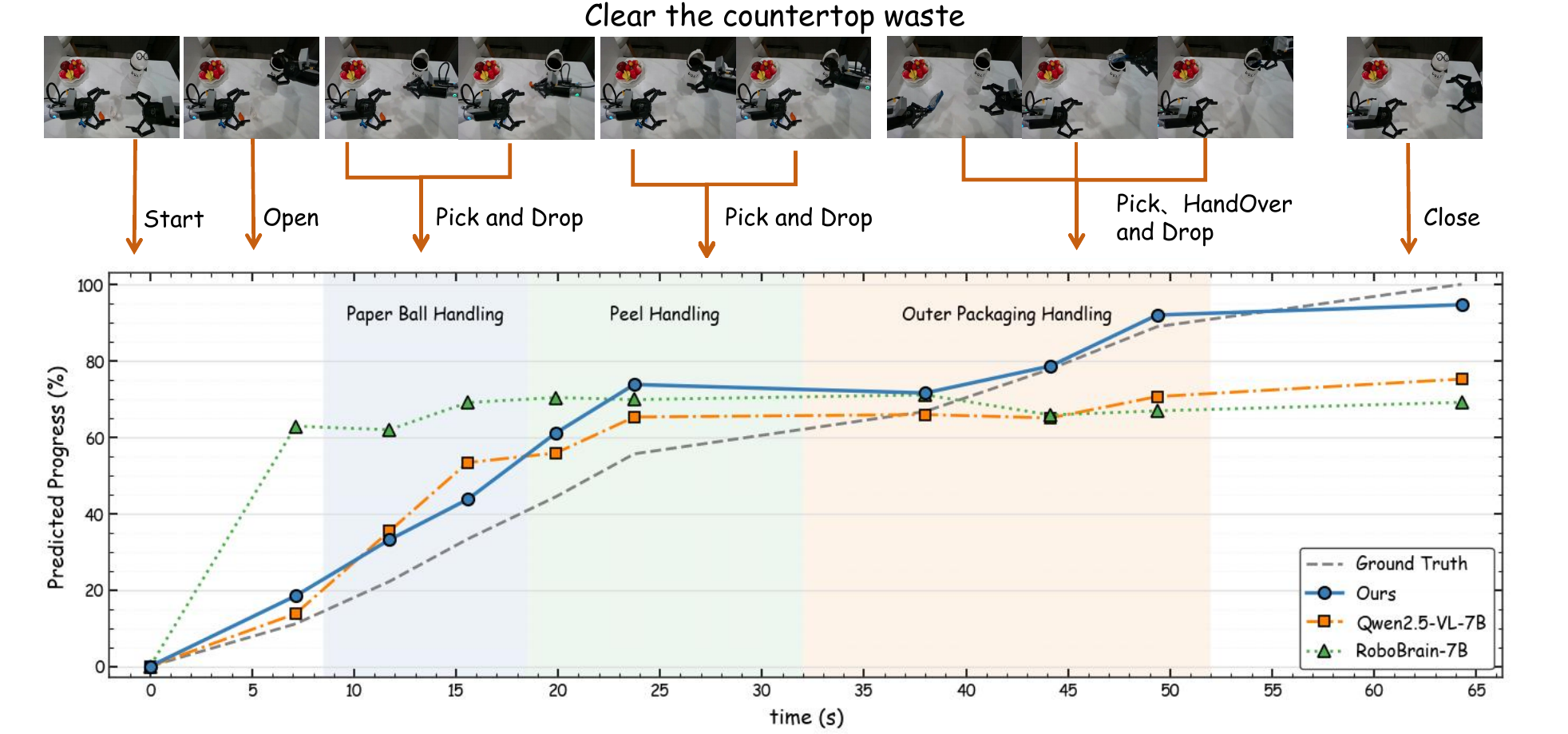}
    \caption{\textbf{Continuous Progress Estimation.} Average predicted progress trajectory over 105 episodes for the ``Clear the countertop waste'' task, comparing temporal state alignment against baselines.}
    \label{fig:progress_estimation}
\end{figure}

To isolate the impact of temporal context and state representations on progress estimation, we conduct an ablation study over three input variables: the initial state image $I_{init}$, the process video sequence $V_{seq}$, and the current state image $I_{curr}$. The quantitative results are summarized in Table~\ref{tab:ablation_temporal}.

Using only the current state $I_{curr}$ yields the highest estimation error, with an Average MAE of 59.50, indicating that a static snapshot lacks sufficient causal context for accurate progress estimation. Replacing it with the temporal sequence $V_{seq}$ reduces the Average MAE to 36.61, but performance remains limited across most tasks, suggesting that temporal information alone is insufficient without explicit reference states.

Prior work~\cite{chen2026egoplan, chen2025grpo} has shown the benefits of incorporating $I_{curr}$ in video-centric planning. Consistent with this observation, adding either $I_{init}$ or $I_{curr}$ to $V_{seq}$ further improves estimation performance. Specifically, $V_{seq}+I_{curr}$ achieves the lowest MAE on RoboTwin (40.24), while $I_{init}+V_{seq}$ performs best on AgiBot (24.94). Our final PRIMO R1 adopts all three modalities. Although dual-modality combinations perform competitively on shorter-horizon datasets such as AgiBot, incorporating both boundary states yields a clear advantage on the long-horizon Behavior dataset, reducing MAE to 22.73 and improving Acc@10 to 31.83. Conversely, the boundary-state paradigm ($I_{init}+I_{curr}$) used in prior work, such as SCIZOR~\cite{zhang2026scizor} suffers without video input, increasing the Behavior MAE from 22.73 to 49.59. While the gap is smaller on RoboTwin (42.16 vs. 45.93), the substantial degradation on long-horizon tasks confirms that explicit video modeling is equally essential.

These results highlight a fundamental representation trade-off. Video features effectively capture temporal dynamics but sacrifice fine-grained spatial information needed for accurate state comparison. Explicitly incorporating both $I_{init}$ and $I_{curr}$ anchors the temporal evolution encoded by $V_{seq}$, providing complementary spatial references for robust progress estimation.


\begin{figure}[t]
    \vspace{-8pt}
    \centering
    \includegraphics[width=0.9\linewidth]{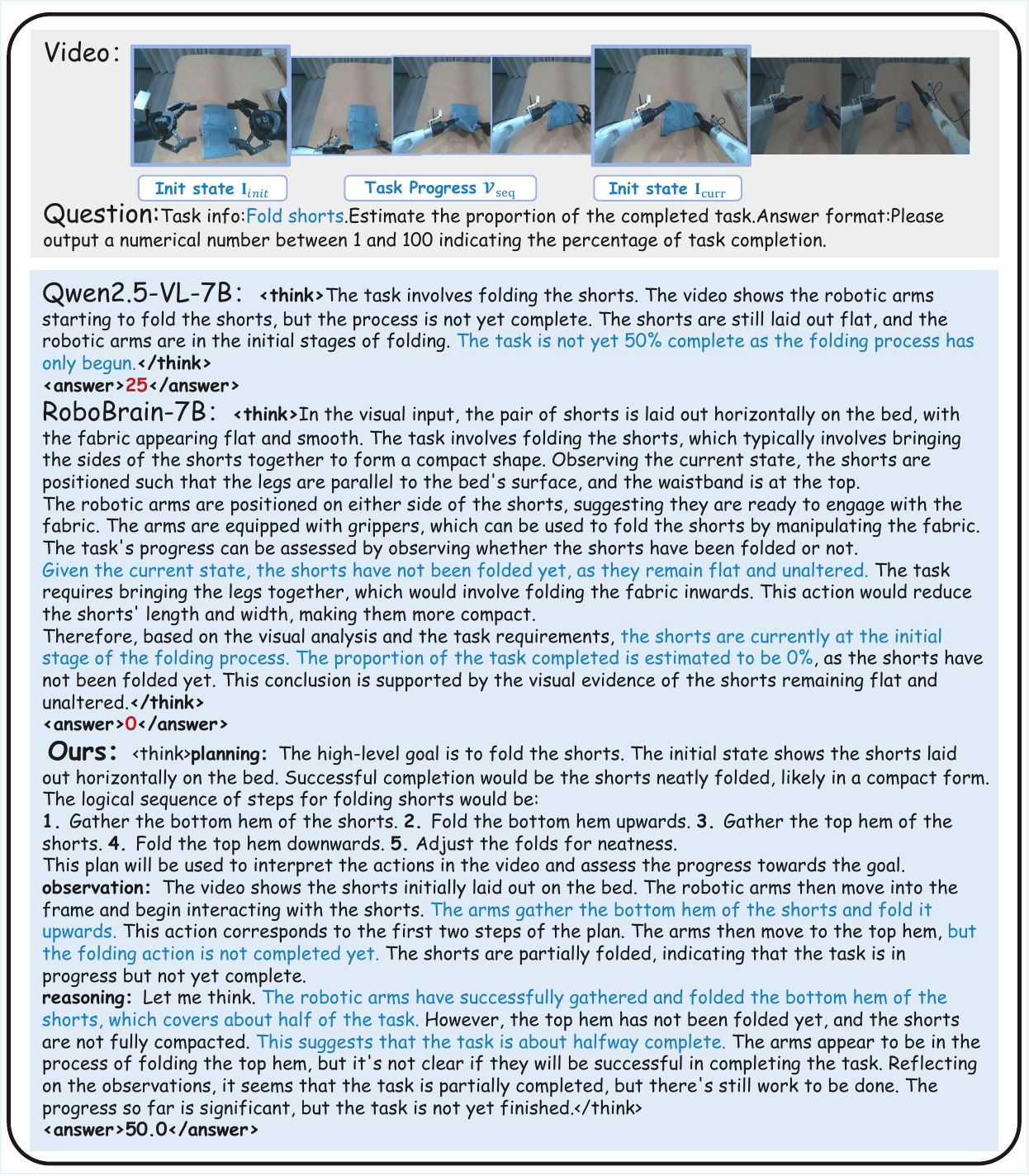}
    \caption{\textbf{Qualitative Comparison of Reasoning Processes.} This case study illustrates the generated reasoning chains of Qwen2.5-VL-7B, RoboBrain-7B and our PRIMO R1 during the Fold shorts task in AgiBot environment.}
    \label{fig:case_study_reasoning}
    \vspace{-8pt}
\end{figure}

\subsection{Case Study}

To evaluate continuous tracking capabilities in long-horizon scenarios, Figure~\ref{fig:progress_estimation} plots the predicted progress trajectory for the composite task ``Clear the countertop waste'', mapping the average outputs across 105 episodes. The evaluation isolates the relationship between the predicted progress variable and the ground-truth temporal execution sequence. The baseline RoboBrain-7B demonstrates a decoupling from actual physical states; its prediction overshoots during the initial 0-10s phase and subsequently loses sensitivity to further temporal advancement. Qwen2.5-VL-7B tracks the initial sub-stages accurately, but its prediction variable plateaus near 60\%-70\% during the latter half, failing to map visual convergence to the final task state. Our model maintains a monotonically increasing trajectory that correlates linearly with the ground truth across discrete sub-task transitions. In the terminal phase, it accurately maps the visual state change of the final action to a progress metric approaching 100\%, verifying a stable alignment between long-range temporal sequences and progress estimation.

Figure~\ref{fig:case_study_reasoning} details the structural decomposition of the explicit reasoning chain generated by the model for a ``Fold shorts'' video. Baselines struggle with fine-grained state tracking: RoboBrain-7B overlooks ongoing dynamic manipulations (0\% progress), while Qwen2.5-VL-7B lacks a structured evaluation metric (25\% progress). Conversely, our PRIMO R1 generates an explicit reasoning chain via three modules. The \textit{Planning} module establishes a reference topology by breaking down the high-level semantic goal into a linear five-step execution plan (Gather bottom hem $\rightarrow$ Fold upwards $\rightarrow$ Gather top hem $\rightarrow$ Fold downwards $\rightarrow$ Adjust). The \textit{Observation} module discretizes the continuous visual input, extracting specific dynamic variables and verifying part-level object state changes (e.g., isolating the state of the bottom hem from the top hem). Finally, the \textit{Reasoning} module executes state alignment by mapping the extracted visual primitives against the planned execution topology. It identifies precise execution boundaries. Specifically, confirming the successful manipulation of the bottom hem while explicitly verifying the incomplete status of the top hem, which acts as a structural constraint for quantitative evaluation. The final numerical prediction (50.0\%) is formulated by calculating the ratio of the verified execution steps against the complete reference plan. Since Inference latency and real-time performance are critical for robotic manipulation, we also provide an analysis and comparison of reasoning chain lengths and inference times in Appendix~\ref{subsec:appendix_quantitative_efficiency}.

\section{Conclusion}

In this work, we introduced PRIMO R1, a 7B framework that transforms video MLLMs into active critics for robotic process supervision via outcome-based reinforcement learning (GRPO). By explicitly anchoring temporal sequences between initial and current state images and incentivizing Chain-of-Thought generation, our approach mitigates spatial detail dilution and enables rigorous temporal reasoning. Furthermore, conditioning this reasoning process on diverse natural language task goals explicitly exploits the language generalization capabilities of foundational LLMs. Experimental results across simulation and real-world humanoid domains demonstrate that PRIMO R1 achieves state-of-the-art performance, empirically establishing that optimizing continuous progress tracking intrinsically constructs the prerequisite representations for zero-shot discrete failure detection, suggesting a pathway toward deriving reward signals essential for future autonomous policy learning in long-horizon manipulation.

\section*{Acknowledgements}

This work was supported in part by the Shanghai Magnolia Talent Program Pujiang Project under Grant No. 25PJA076.


%
%
\clearpage
\setlength{\textfloatsep}{20pt plus 2pt minus 4pt}
\setlength{\floatsep}{12pt plus 2pt minus 2pt}
\setlength{\intextsep}{12pt plus 2pt minus 2pt}
\bibliographystyle{splncs04}
\bibliography{main}

\newpage

\appendix
\section*{Appendix}
\label{sec:appendix_start}



\section{Benchmark and Datasets Details}
\label{sec:appendix_datasets}

\subsection{Dataset Construction Methodology}

To build a comprehensive and diverse benchmark for embodied agents, we construct our dataset by aggregating multi-source data from both high-fidelity simulations and real-world humanoid robot manipulation. The dataset composition covers a wide range of complexity, from atomic actions to long-horizon composite tasks.

\vspace{1mm}
\noindent\textbf{Simulation Data Collection.}
The simulation data is derived from two primary high-fidelity sources: BEHAVIOR-1k and RoboTwin. 
\begin{itemize}
    \item \textbf{BEHAVIOR-1k:} We source data from the 2025 BEHAVIOR Challenge~\cite{li2023behavior}. To enrich the semantic annotations, we employ Large Language Models (LLMs) to convert the original BDDL-based annotations into natural language captions, followed by timestamp-based segmentation to derive fine-grained sub-tasks (ranging from 4 to 76 steps).
    \item \textbf{RoboTwin:} We adopt the code generation methodology proposed in Hycodepolicy~\cite{liu2025hycodepolicy} for the RoboTwin~\cite{chen2025robotwin} simulator. By automatically injecting sub-task and timestamp markers into the generated code, we efficiently synthesize and split the data into trajectory segments.
\end{itemize}

\vspace{1mm}
\noindent\textbf{Real-World Data Collection.}
To capture the complexity of physical environments and bridge the sim-to-real gap, we incorporate real-world data from two distinct platforms, serving different phases of our post-training and evaluation paradigm:
\begin{itemize}
    \item \textbf{AgiBot (Training \& In-Domain):} Serving as the primary real-world component of our training corpus, we utilize the AgiBot dataset~\cite{bu2025agibot}. We process the raw real-world teleoperation data by utilizing timestamps to segment task progress and extract sub-task demonstrations.
    \item \textbf{Real Humanoid (OOD Evaluation):} To construct a stringent Cross-Environment generalization benchmark, we collect a supplementary real-world, multi-task dataset via teleoperation of the Kuavo 4 Pro full-size humanoid robot from LejuRobotics Technology Co., Ltd. This dataset encompasses multi-scenario and multi-type operations targeting robot manipulation, locomotion, and interaction tasks. It is designed to support scalable robot learning in diverse unstructured physical environments, including hotel services, manufacturing factories, fast-moving consumer goods (FMCG) scenarios, and automotive assembly lines.
\end{itemize}

\subsection{Dataset Statistics and Distribution}

Table~\ref{tab:dataset_stats} summarizes the statistics of the constructed dataset, detailing the task distribution, video counts, and the scale of processed trajectory segments across different domains. Figure~\ref{fig:dataset_dist} visualizes the data distribution splits across the SFT phase, RL phase, and the PRIMO Bench evaluation sets.

\begin{table}[h]
\centering
\caption{Statistics of the constructed dataset. The table details the number of tasks, raw video demonstrations, sub-task complexity, and the final volume of processed data samples for training and evaluation.}
\label{tab:dataset_stats}
\resizebox{\textwidth}{!}{%
\begin{tabular}{lcccccc}
\toprule
\textbf{Dataset} & \textbf{Source} & \textbf{\# Tasks} & \textbf{Split (Train / Test)} & \textbf{Raw Videos} & \textbf{Sub-tasks (Min-Max)} & \textbf{Processed Samples} \\
\midrule
\textbf{AgiBot} & AgiBot World~\cite{bu2025agibot} & 36 & 30 / 6 & 7,576 & 1 - 16 & 48,276 \\
\textbf{Behavior} & BEHAVIOR-1k~\cite{li2023behavior} & 50 & 40 / 10 & 9,992 & 4 - 76 & 235,441 \\
\textbf{RoboTwin} & RoboTwin~\cite{chen2025robotwin} & 49 & 35 / 14 & 24,500 & 1 - 9 & 71,708 \\
\textbf{Real Humanoid} & Real World (KUAVO-MY) & 7 & - / 7 & 2,800 & 2 - 5 & 2,800$^{*}$ \\
\midrule
\textbf{Total} & - & \textbf{150} & - & \textbf{32,868} & - & \textbf{326,453} \\
\bottomrule
\multicolumn{7}{l}{\footnotesize{*For Real Humanoid, processed samples represent the validation set count utilized for evaluation.}}
\end{tabular}%
}
\end{table}

\subsection{Other Benchmark and Datasets Details}

\noindent\textbf{RoboFail Benchmark:} Curated and annotated by Cosmos-Reason1~\cite{azzolini2025cosmos}, this benchmark originates from the RoboFail dataset introduced in REFLECT~\cite{liu2023reflect}. It comprises an evaluation split of 100 examples focusing on harder ``action affordance'' and ``task completion verification'' scenarios. The hardness of these samples is dictated by the necessity for highly observant perception or comprehensive temporal context processing, requiring models to identify physical constraints blocking the follow-through for an action and to reason about nuanced questions.

\vspace{2mm}
\noindent Beyond the primary benchmarks used for progress estimation, we incorporate several multimodal datasets during the training process to enhance the model's capabilities in task planning, temporal reasoning, and scene understanding. These datasets provide the diverse semantic and structural supervision necessary for the transition from a passive observer to an active critic.

\begin{itemize}
    \item \textbf{ShareRobot Dataset} \cite{ji2025robobrain}: A high-quality heterogeneous dataset featuring multi-dimensional annotations including object affordance and end-effector trajectories. In our framework, we exclusively utilize the \textbf{task planning} data, which includes high-quality heterogeneous labels used to enhance the model's abstract reasoning and goal decomposition capabilities.
    
    \item \textbf{EgoPlan-Bench} \cite{chen2026egoplan}: A comprehensive benchmark designed to evaluate the planning abilities of MLLMs in real-world scenarios from an egocentric perspective. It focuses on human-level planning through diverse action plans and intricate visual observations to mirroring human perception.
    
    \item \textbf{RoboVQA} \cite{sermanet2024robovqa}: A large-scale, diverse dataset containing video-text pairs for robotics-focused visual question answering. It supports the development of models capable of grounded, high-level reasoning across long-horizon tasks and multiple embodiments.
    
    \item \textbf{Perception Test} \cite{patraucean2023perception}: A diagnostic benchmark that evaluates perception and reasoning skills—such as memory, abstraction, physics, and semantics—using real-world videos densely annotated with multiple-choice and grounded video question-answers.
    
    \item \textbf{STAR} \cite{li2025star}: A large-scale dataset for scene graph generation in high-resolution satellite imagery. It promotes geospatial scenario understanding by requiring long-range contextual reasoning to mine triplets of subjects, relationships, and objects.
    
    \item \textbf{NExT-QA} \cite{xiao2021next}: A video question-answering benchmark designed to advance video understanding beyond shallow descriptions toward explaining temporal actions. It specifically targets causal action reasoning, temporal action reasoning, and common scene comprehension.
\end{itemize}

\begin{figure}[t]
    \centering
    \begin{minipage}{0.48\textwidth}
        \centering
        \includegraphics[width=\linewidth]{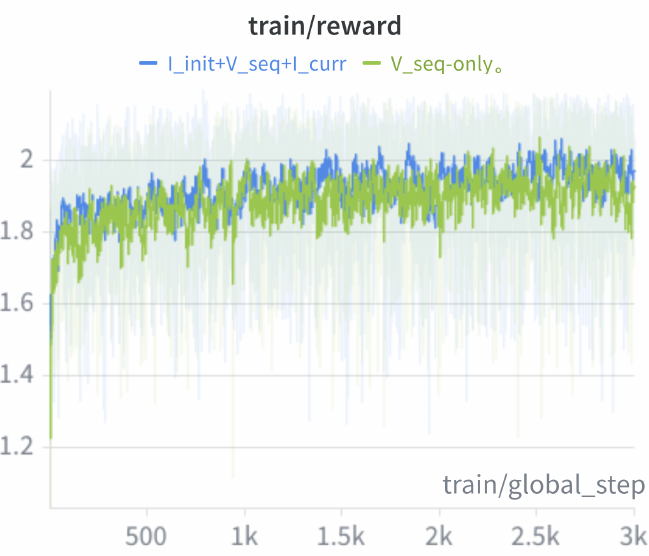}
    \end{minipage}\hfill
    \begin{minipage}{0.48\textwidth}
        \centering
        \includegraphics[width=\linewidth]{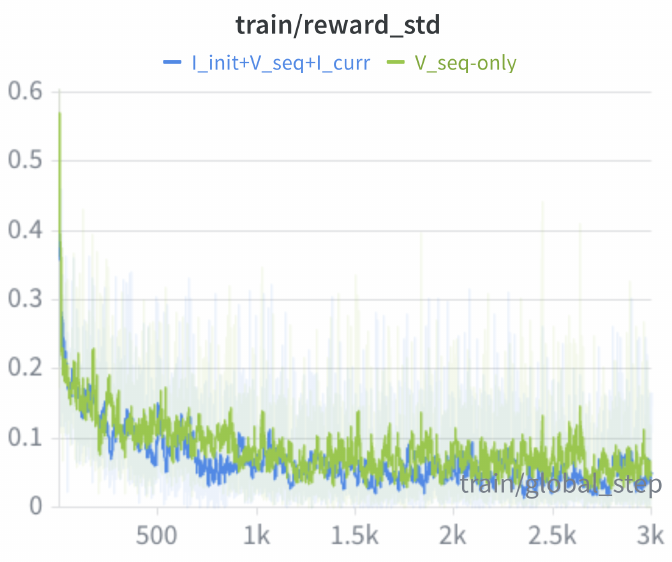}
    \end{minipage}
    \caption{\textbf{Training Dynamics of GRPO Across Input Modalities.} (Left) Average reward curves during GRPO training. The triad modality configuration ($I_{init} + V_{seq} + I_{curr}$) accelerates convergence and achieves higher optimal rewards compared to the pure temporal baseline. (Right) Standard deviation of rewards. The integration of static boundary states explicitly limits the policy variance, ensuring a more stable optimization trajectory.}
    \label{fig:grpo_dynamics}
\end{figure}

\section{Training Dynamics of GRPO Across Input Modalities}
\label{sec:appendix_grpo_dynamics}

To further validate the necessity of explicit boundary state modeling in reinforcement learning, we analyze the training dynamics of Group Relative Policy Optimization (GRPO) under two distinct input modality conditions: the pure temporal sequence ($V_{seq}$) and our proposed triad configuration ($I_{init} + V_{seq} + I_{curr}$). The evaluation metrics focus on the average reward convergence and the standard deviation of rewards across training steps.

As illustrated in Figure~\ref{fig:grpo_dynamics}, the inclusion of static boundary images fundamentally alters the optimization landscape. 

\textbf{Reward Convergence:} Figure~\ref{fig:grpo_dynamics}(a) demonstrates that the triad configuration achieves faster convergence and a higher final reward plateau compared to the $V_{seq}$-only baseline. The explicit inclusion of $I_{init}$ and $I_{curr}$ provides the policy model with structural spatial anchors, mitigating the credit assignment difficulties typically associated with evaluating long-horizon temporal features.

\textbf{Training Stability:} Figure~\ref{fig:grpo_dynamics}(b) plots the standard deviation of the rewards within the GRPO sampling groups. The $V_{seq}$-only model exhibits higher variance, indicating instability in policy updates due to the lack of static alignment constraints. In contrast, explicitly tracking the transition from $I_{init}$ to $I_{curr}$ confines the policy search space, significantly reducing the reward variance and yielding a more stable fine-tuning process.

\begin{figure}[h]
    \vspace{-8pt}
    \centering
    \includegraphics[width=0.9\linewidth]{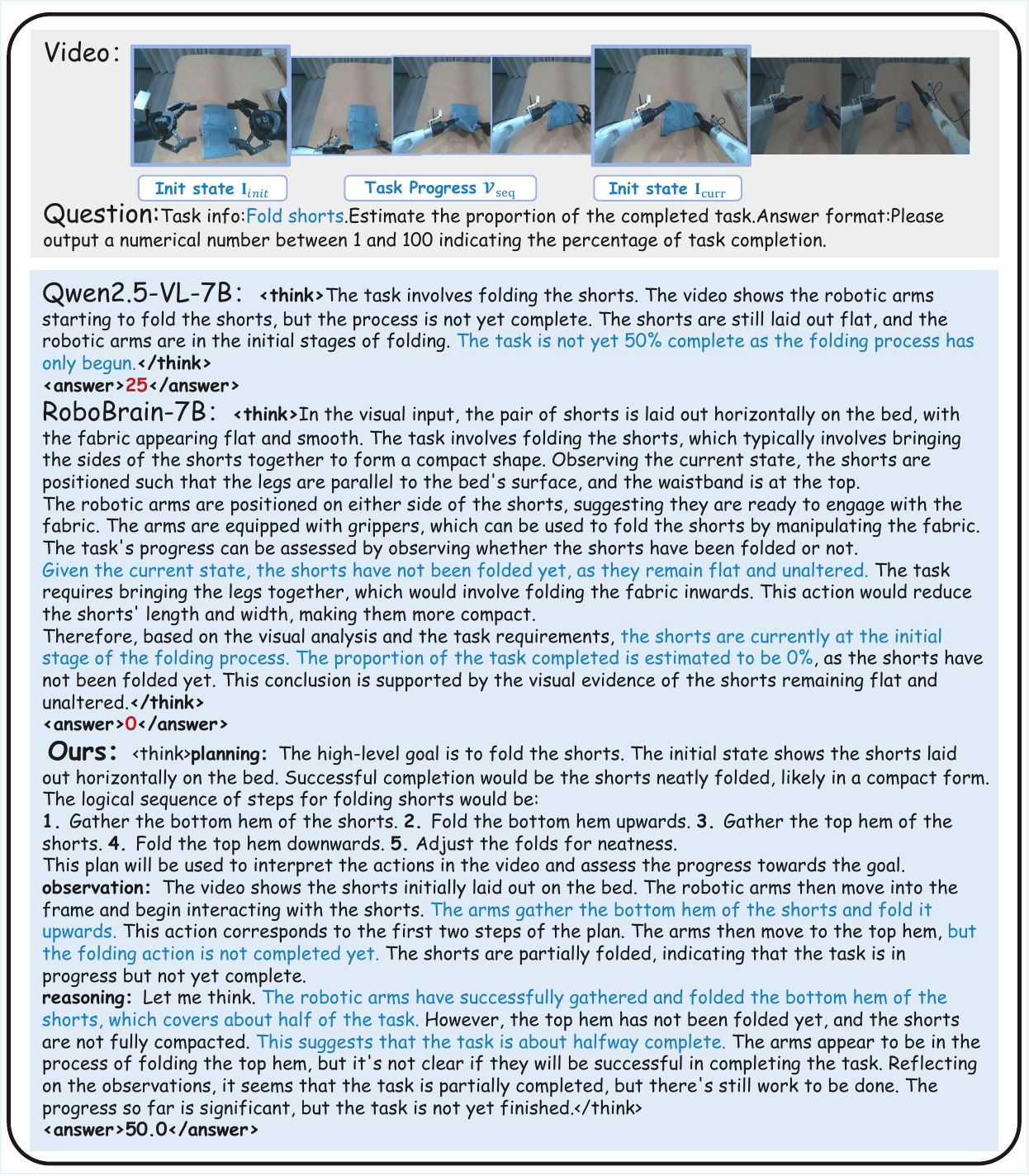}
    \caption{\textbf{Qualitative Comparison of Reasoning Processes.} This case study illustrates the generated reasoning chains of Qwen2.5-VL-7B, RoboBrain-7B and our PRIMO R1 during the Fold shorts task in AgiBot environment.}
    \label{fig:qualitative_reasoning_case2}
    \vspace{-15pt}
\end{figure}

\section{Detailed Analysis of Reasoning Processes}
\label{sec:appendix_case}

\subsection{Qualitative Comparison of Reasoning Processes}
\label{subsec:appendix_qualitative_case}
Figure~\ref{fig:qualitative_reasoning_case2} compares the reasoning processes during the ``Fold shorts'' task. Baselines struggle with fine-grained state tracking: RoboBrain-7B overlooks ongoing dynamic manipulations (0\% progress), while Qwen2.5-VL-7B lacks a structured evaluation metric (25\% progress). Conversely, our PRIMO R1 generates an explicit reasoning chain via three modules. The \textit{planning} module decomposes the semantic goal into a five-step reference topology (Gather bottom hem $\rightarrow$ Fold upwards $\rightarrow$ Gather top hem $\rightarrow$ Fold downwards $\rightarrow$ Adjust). The \textit{observation} module discretizes visual inputs, verifying the execution of the first two steps. Finally, the \textit{reasoning} module performs state alignment by mapping visual primitives against the planned topology. By confirming the successful manipulation of the bottom hem alongside the incomplete top hem, PRIMO R1 calculates the ratio of verified steps to formulate a precise and interpretable prediction of 50.0\%.

\begin{figure}[t]
    \vspace{-5pt}
    \centering
    \includegraphics[width=1\linewidth]{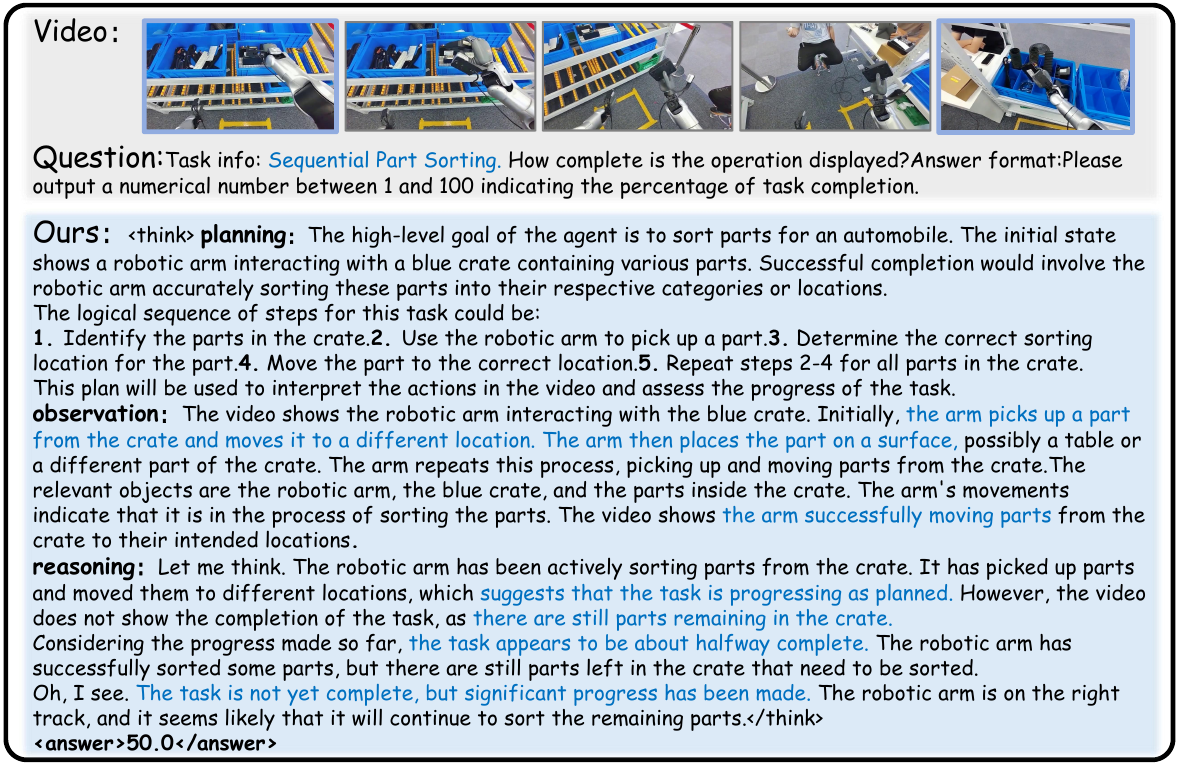}
    \caption{An example of our PRIMO R1’s reasoning out put on Real Humanoid environment.}
    \label{fig:casesappen}
    \vspace{-15pt}
\end{figure}

Figure~\ref{fig:casesappen} further demonstrates PRIMO R1's reasoning capability in a physical Real Humanoid environment during a ``Sequential Part Sorting'' task. Unlike the rigid sequential constraints of the previous example, this task requires assessing iterative cyclic actions. The \textit{planning} module formulates a loop-based execution strategy (Identify $\rightarrow$ Pick $\rightarrow$ Locate $\rightarrow$ Move $\rightarrow$ Repeat). The \textit{observation} module continuously monitors this iterative process, successfully recognizing the recurring pick-and-place actions alongside the changing state of the source crate. During the \textit{reasoning} phase, the model synthesizes these visual cues to evaluate global progress. By logically recognizing that multiple parts have been successfully sorted while others visibly remain in the crate, PRIMO R1 deduces the partial completion state, outputting a grounded 50.0\% progress estimation without requiring explicit part counting.

\subsection{Quantitative Analysis: Chain Length and Inference Latency}
\label{subsec:appendix_quantitative_efficiency}

To comprehensively evaluate the feasibility of models for real-time robotic operations, we analyze the inference efficiency (Table~\ref{tab:inference_efficiency}) in conjunction with the task progress estimation performance (MRA) from our main results. Relying solely on raw latency or token count is insufficient, as an optimal model must strike a balance between reasoning efficiency and predictive accuracy. 

By jointly considering MRA and inference costs, a clear performance-efficiency trade-off emerges. Notably, Cosmos-Reasoning 7B demonstrates severe inefficiency; despite generating the most verbose reasoning chains (averaging 1109.75 tokens) and incurring the highest latency (1.30s), it yields the lowest average MRA (66.52) among the evaluated reasoning MLLMs. This indicates that its prolonged generation fails to translate into effective structural reasoning. Conversely, while Robobrain 7B is the fastest (0.50s) with the shortest token lengths (242.3), it sacrifices substantial accuracy, trailing our model by nearly 15\% in average MRA. 

Our PRIMO R1 achieves an optimal balance. With an average inference latency of 0.62s and a concise reasoning chain of 359.18 tokens, its computational overhead remains strictly competitive with fast baselines like Video R1 7B. However, PRIMO R1 delivers a massive leap in accuracy, achieving an average MRA of 82.90. When evaluating efficiency as the accuracy gained per unit of inference time or token, PRIMO R1 stands out as the most cost-effective solution, proving highly effective and responsive for real-time robotic manipulation tasks.

\begin{table}
\centering
\caption{\textbf{Comparison of Inference Efficiency.} We report the inference latency (\textit{time} in seconds) and reasoning chain length (\textit{token} count) across four distinct environments. This quantitative analysis evaluates the feasibility and efficiency of models for real-time robotic operations.}
\label{tab:inference_efficiency}
\resizebox{0.9\textwidth}{!}{%
\begin{tabular}{l|cc|cc|cc|cc|cc}
\toprule
\multirow{2}{*}{\textbf{Model}} & \multicolumn{2}{c|}{\textbf{AgiBot}} & \multicolumn{2}{c|}{\textbf{Behavior}} & \multicolumn{2}{c|}{\textbf{RoboTwin}} & \multicolumn{2}{c|}{\textbf{Humanoid}} & \multicolumn{2}{c}{\textbf{Average}} \\
 & time   & token   & time   & token   & time   & token   & time   & token   & time   & token   \\
\midrule
Video R1 7B & 0.53 & 381.3 & 0.58 & 364.9 & 0.56 & 375.5 & 0.50 & 383.74 & 0.54 & 376.36 \\
Robobrain 7B & 0.50 & 247.32 & 0.47 & 201.3 & 0.53 & 262.74 & 0.48 & 257.84 & 0.50 & 242.3 \\
Cosmos-Reasoning 7B & 1.42 &1394.40 & 1.30 & 868.92 & 1.13 & 795.70 & 1.36 & 1379.96 & 1.30 & 1109.75 \\
\midrule
\rowcolor{graybg} \textbf{PRIMO R1 (Ours)} & 0.64 & 360.16 & 0.61 & 363.9 & 0.60 & 350.98 & 0.61 & 361.66 & 0.62 & 359.18 \\
\bottomrule
\end{tabular}%
}
\vspace{-15pt}
\end{table}

\section{Cross-Benchmark Generalization on ProgressLM}
\label{sec:appendix_progresslm}

To assess whether PRIMO R1 generalizes beyond PRIMO Bench, we evaluate it on the external ProgressLM benchmark~\cite{zhang2026progresslm} without any additional fine-tuning. This benchmark differs from our setting in two respects. First, it supplies a densely annotated reference sequence (e.g., paired demonstration frames labeled with $0\%$, $50\%$, and $100\%$ progress) and reduces progress estimation to matching the current state against these references. Second, the ProgressLM models are trained directly on this benchmark, whereas PRIMO R1 is applied zero-shot and estimates progress without any reference demonstration. The comparison is therefore not strictly head-to-head; we report it to characterize transfer to an unseen evaluation protocol.

Table~\ref{tab:progresslm} reports results using Normalized Score Error (NSE $\downarrow$), Progress Rank Correlation (PRC $\uparrow$), and Answerable False Rejection Rate (AFRR $\downarrow$). Two observations stand out. First, among models of comparable scale, PRIMO R1 attains the lowest error and the highest rank correlation; it also surpasses the much larger Intern3.5-VL-38B on the micro and macro averages despite using fewer parameters and no reference frames. Second, PRIMO R1 maintains a consistently low AFRR across both the vision-based and text-based settings (at most $1.0$), while several baselines either reject a large fraction of answerable cases (e.g., Intern3.5-VL-38B at $38.3$ micro) or achieve a low rejection rate only by collapsing on the other two metrics. The reference-trained ProgressLM-3B-SFT obtains a lower NSE and a higher PRC, which is expected given that it is optimized on this benchmark with explicit reference supervision; even so, PRIMO R1 retains a clear advantage on AFRR. These results indicate that explicit process reasoning transfers across benchmarks and yields stable behavior on answerable queries, rather than overfitting to a single evaluation format.

\begin{table}[h]
\centering
\caption{\textbf{Cross-benchmark evaluation on the external ProgressLM benchmark}~\cite{zhang2026progresslm}. Each cell reports (NSE $\downarrow$ / PRC $\uparrow$ / AFRR $\downarrow$). PRIMO R1 is evaluated zero-shot without reference demonstrations, whereas the ProgressLM models are trained on this benchmark with reference supervision. Best AFRR in each column is in \textbf{bold}.}
\label{tab:progresslm}
\setlength{\tabcolsep}{4pt}
\resizebox{\textwidth}{!}{%
\begin{tabular}{l|cccc}
\toprule
\textbf{Model} & \textbf{Vision Demo} & \textbf{Text Demo} & \textbf{Micro Avg} & \textbf{Macro Avg} \\
\midrule
Qwen2.5-VL-7B & 34.0 / 33.7 / 28.3 & 39.1 / 20.5 / \textbf{0.0} & 35.0 / 31.0 / 22.5 & 36.5 / 27.1 / 14.2 \\
Qwen2.5-VL-32B & 38.9 / 41.5 / \textbf{0.0} & 42.6 / 30.0 / \textbf{0.0} & 39.7 / 39.2 / \textbf{0.0} & 40.8 / 35.8 / \textbf{0.0} \\
Intern3.5-VL-14B & 65.2 / -22.3 / 0.2 & 39.5 / 10.3 / 6.8 & 60.0 / -15.6 / 1.5 & 52.3 / -6.0 / 3.5 \\
Intern3.5-VL-38B & 35.2 / 56.7 / 37.1 & 26.5 / 23.3 / 43.1 & 33.4 / 49.9 / 38.3 & 30.8 / 40.0 / 40.1 \\
ProgressLM-3B-SFT & 19.0 / 72.4 / 6.5 & 29.1 / 46.3 / 9.2 & 21.1 / 67.0 / 7.0 & 24.0 / 59.3 / 7.8 \\
\midrule
\rowcolor{graybg} \textbf{PRIMO R1 (Ours)} & 32.9 / 53.8 / \textbf{0.0} & 32.2 / 32.8 / 1.0 & 32.8 / 49.5 / 0.2 & 32.6 / 43.3 / 0.5 \\
\bottomrule
\end{tabular}%
}
\end{table}

\section{Inference and RL Training Prompt}
\label{sec:appendix_rl_prompt}

In this section, we detail the comprehensive prompt structure used to elicit Chain-of-Thought (CoT) reasoning for Process Reasoning Induced Monitoring (\textbf{PRIMO R1}). This prompt is designed to enforce a rigorous internal thought process across three specific dimensions: planning, observation, and reasoning.
\section{Inference and RL Training Prompt}
\label{sec:appendix_rl_prompt}

\begin{figure}[t]
\centering
\begin{tcolorbox}[
    enhanced,
    colback=gray!5!white,          
    colframe=gray!60!black,        
    boxrule=0.5pt,                 
    arc=2mm,                       
    auto outer arc,
    title=\textbf{Prompt Template: Task Progress Estimation}, 
    coltitle=black,                
    colbacktitle=gray!15!white,    
    fontupper=\small\ttfamily,     
    left=6pt, right=6pt, top=6pt, bottom=6pt, 
    drop shadow=gray!25!white      
]

\textbf{Task info:}\textcolor{blue!70!black}{\{task\_info\}}\\
\textbf{Init Scene:}\textcolor{blue!70!black}{\{init\_scene\_text\}} (Option)\\
\textbf{Question:}Estimate the completion percentage of the task. (0-100\%)\\
\textbf{Answer format:}Please output a numerical value between 1 and 100 indicating the percentage of task completion.
\end{tcolorbox}
\caption{The prompt template used for querying the Video MLLM to estimate task progress. Dynamic variables are highlighted in blue. The displayed question \textit{``Estimate the completion percentage of the task. (0-100\%)''} serves as a representative example. To ensure prompt robustness and improve instruction generalization, we synthesized 100 distinct question variations for querying task progress. The comprehensive list of these variations is provided in Appendix~\ref{sec:appendix_prompts}.}
\label{fig:prompt_template}
\end{figure}

The System Prompt defines the foundational persona of \textbf{PRIMO R1} and establishes the multi-modal input processing protocol. It explicitly instructs the model to synthesize the initial state, the temporal video sequence, and the current state to ensure a grounded understanding of the task progression.

\begin{tcolorbox}[
    breakable,
    enhanced,
    colback=gray!2!white,
    colframe=gray!50!black,
    title=\textbf{User Prompt for Embodied Procedure Reasoning},
    colbacktitle=gray!20!white,
    coltitle=black,
    fontupper=\small\ttfamily,
    arc=1mm,
    boxrule=0.7pt,
    left=8pt, right=8pt, top=8pt, bottom=8pt
]

A conversation between User and Assistant. The Assistant is an expert AI specializing in embodied procedure and event reasoning based on visual input. 

\vspace{0.5em}
You will be provided with three types of visual information:\\
(1) Initial State - an image showing the starting condition,\\
(2) Video - capturing the procedure from Initial State to Current State,\\
(3) Current State - an image showing the ending condition. 

\vspace{0.5em}
You must analyze all three inputs together to understand the complete task progression and answer the question. 

\vspace{0.5em}
The assistant must strictly follow a specific thought process and output format. The reasoning process is enclosed within <think> </think> tags, and the final answer is within <answer> </answer> tags. 

\vspace{0.5em}
The <think> block must contain three ordered subsections: <planning>, <observation>, and <reasoning>. 

\vspace{0.5em}
The <answer> block must contain only the final output required by the question type and no other commentary.
\end{tcolorbox}

\begin{tcolorbox}[
    breakable,
    enhanced,
    colback=gray!2!white,
    colframe=gray!50!black,
    title=\textbf{User Prompt for Embodied Procedure Reasoning},
    colbacktitle=gray!20!white,
    coltitle=black,
    fontupper=\small\ttfamily,
    arc=1mm,
    boxrule=0.7pt,
    left=8pt, right=8pt, top=8pt, bottom=8pt
]

QUESTION:\\
\textcolor{blue!70!black}{\{Question\}}

\vspace{0.5em}
QUESTION TYPE:\\
\textcolor{blue!70!black}{\{question\_type\}}

\vspace{1em}
Analyze the provided visual data and reason about the ongoing task. 

\vspace{0.5em}
Please think about this question as if you were a human pondering deeply. Provide your detailed reasoning between the <think> and </think> tags, following the subsections <planning>, <observation>, and <reasoning>. Then give your final answer between the <answer> and </answer> tags.

\vspace{0.5em}
Below is the required template:

\vspace{0.5em}
<think>\\
<planning>\\
Identify the high-level goal of the agent, what is the initial state? What does successful completion look like?\\
Break down the high-level goal into a logical sequence of canonical steps. This serves as your mental plan for interpreting the task.\\
Use this plan to interpret actions, map observed behaviors to steps, assess progress, detect anomalies, and predict what happens next.\\
</planning>

\vspace{0.5em}
<observation>\\
View the video as a temporal sequence of actions contributing to the procedure.\\
Objectively describe what is occurring in the current moment, noting evidence of progress or state changes.\\
Identify fine-grained actions and explain how they move the task forward.\\
List relevant objects, tools, and environmental context, emphasizing functional states and transformations.\\
Note cues—repetition, transitions, or completion indicators—that situate the action in the procedural script.\\
</observation>

\vspace{0.5em}
<reasoning>\\
Think through the question as a human would, engage in an internal dialogue using expressions such as 'let me think', 'wait', 'hmm', 'oh, I see', 'let's break it down', etc.\\
Connect observations to the procedural plan to determine which step is being executed, progress, correctness, or anomalies.\\
Reflect on assumptions, verify interpretations, and, if appropriate, predict the agent's next likely action.\\
Synthesize understanding of what the agent is doing, how it fits into the broader task, and whether the process seems successful.\\
You are encouraged to include self-reflection or verification in your reasoning process.\\
</reasoning>\\
</think>

\vspace{0.5em}
<answer>\\
Final answer here — strictly follow the \textcolor{blue!70!black}{\{question\_type\}} output format and include no extra commentary.
</answer>

\end{tcolorbox}

To ensure the model outputs are verifiable and parsable for reinforcement learning rewards, we enforce strict output constraints based on the task category. Table~\ref{tab:type_template} lists the specific instructions injected into the \texttt{\{question\_type\}} variable.

\begin{table}[ht]
\centering
\caption{Question Type Instructions (\texttt{TYPE\_TEMPLATE}).}
\label{tab:type_template}
\small
\begin{tabular}{@{}ll@{}}
\toprule
\textbf{Category} & \textbf{Appended Instruction within <answer> Block} \\ \midrule
Multiple Choice   & Please provide only the single option letter (e.g., A, B, C, D, etc.). \\
Numerical         & Please provide the numerical value (e.g., 42 or 3.14). \\
OCR               & Please transcribe text from the image/video clearly. \\
Free-form         & Please provide your text answer directly. \\
Boolean           & Please provide only 'Yes' or 'No'. \\
\textbf{Progress} & \textbf{Please output a numerical number between 1 and 100.} \\ \bottomrule
\end{tabular}
\end{table}

\section{Question Variations for Task Progress Estimation}
\label{sec:appendix_prompts}

To enhance the robustness of \textbf{PRIMO R1} and ensure its generalization across diverse linguistic phrasings, we curated a set of 100 distinct question variations. These prompts range from direct inquiries to context-aware evaluations, preventing the model from over-fitting to a single instruction template. The full list of questions used during training and evaluation is provided below:

\begin{enumerate}
    \item How much of the task has been completed?
    \item What percentage of the task is finished?
    \item How complete is the task in the video?
    \item Estimate the completion percentage of the task.
    \item How far along is the agent in completing the task (in percent)?
    \item To what extent has the task been completed?
    \item Please estimate how much of the task has been done (0-100\%).
    \item What fraction of the task appears to be finished?
    \item How much progress has been made toward completing the task?
    \item Give the approximate percentage of task completion.
    \item Based on the video, what is the task's completion percentage?
    \item Considering the ongoing actions, how complete is the task execution?
    \item From the current progress shown, estimate how much of the task is done.
    \item According to the visual evidence, what is the completion rate of the task?
    \item Based on the observed steps, how far has the task progressed?
    \item Judging from the video, how much of the overall task has been achieved?
    \item Based on the actions shown, estimate the percentage of task completion.
    \item Using the video context, determine how much progress has been made.
    \item According to the current situation, what percent of the task is completed?
    \item What is the estimated completion rate of the task shown in this clip?
    \item Task completion percentage?
    \item Estimate task progress (0-100\%).
    \item Completion rate of the task?
    \item Task progress percentage based on the video?
    \item How much of the task is done (in \%)?
    \item Approximate percent of task completion?
    \item Predicted completion level (0-100)?
    \item What's the completion percentage?
    \item Estimate progress ratio (0\% or 100\%)?
    \item Task progress estimation in percentage?
    \item How complete is the overall procedure in the video?
    \item What's the current progress percentage for this task?
    \item Evaluate the current completion level of the task.
    \item How much has the agent accomplished in this task?
    \item Determine the completion percentage of the process.
    \item Provide an estimate of how much of the task is done.
    \item What's the current progress ratio of the operation?
    \item Estimate how complete the ongoing task is.
    \item What is the approximate progress achieved so far?
    \item Based on the video evidence, how much of the task is finished?
    \item According to the observed actions, what percentage is complete?
    \item How far has the agent advanced in completing the task?
    \item Quantify the level of task completion (0-100\%).
    \item Provide a numeric estimate of task completion.
    \item Indicate how much of the task is completed.
    \item What portion of the task has been done so far?
    \item Compute the completion percentage for the current task.
    \item Estimate the proportion of the completed task.
    \item Evaluate the current progress made toward completion.
    \item How progressed is the task shown in this video?
    \item Based on this clip, what’s the completion percentage?
    \item How much progress has the agent made so far?
    \item Indicate the task completion rate as a percentage.
    \item What’s the estimated completion percentage of the shown task?
    \item Approximately what percentage of the task is complete?
    \item How advanced is the task execution in this clip?
    \item What is the current task progress in numeric terms?
    \item From the visual information, estimate the completion percent.
    \item Provide an approximate completion percentage.
    \item How far along toward completion is the task?
    \item Based on the actions, how complete is the task process?
    \item What is the overall completion rate of this task?
    \item Estimate the progress level of the operation (0-100).
    \item To what degree is the task completed according to the video?
    \item Provide an estimation of the task completion level.
    \item How much work has been completed in the task so far?
    \item How complete is the process illustrated in the video?
    \item What's the approximate task completion ratio?
    \item How much of the procedure has been achieved?
    \item Provide a numerical estimate of progress toward completion.
    \item Based on what's shown, estimate the completion level.
    \item How much of the total work has been finished?
    \item Provide a completion score between 0 and 100.
    \item What is the predicted task completion rate?
    \item Please quantify how much progress the agent has made.
    \item How much of the defined task has already been accomplished?
    \item What's the expected percentage of task completion?
    \item From this video, estimate how much the task has progressed.
    \item How much progress can be observed in the task execution?
    \item What is the level of completion observed?
    \item According to the video, what's the completion score?
    \item How complete is the operation displayed?
    \item Determine the degree of completion (in percentage).
    \item How far toward full completion has the agent progressed?
    \item Report the completion rate inferred from the video.
    \item Provide a completion estimate between 0 and 100 percent.
    \item What is the overall completion percentage observed?
    \item How much of the ongoing task is done so far?
    \item What is the measured completion proportion?
    \item Estimate the current percentage of finished work.
    \item Quantify the extent of completion visible in the video.
    \item How far along is the process in percentage terms?
    \item What percentage of the work has been achieved?
    \item Approximate how complete the shown procedure is.
    \item Indicate how much of the task remains unfinished.
    \item How close to full completion is the task right now?
    \item What percentage of the total task goal has been reached?
    \item How much of the intended activity has been completed?
    \item Give an estimated completion rate (0-100\%).
    \item Estimate the degree of completion based on the given video.
\end{enumerate}

\section{Experimental Setup and Config}
\label{sec:appendix_setup}

We conduct all training experiments on a compute node equipped with 8 NVIDIA A100 (80GB) GPUs. 
During the training phase, to balance computational efficiency and temporal modeling capabilities, we limit the input video sequence to a maximum of 16 frames. 
The frame resolution is configured to $128 \times 28 \times 28$ pixels. Detailed hyperparameters and specific experimental configurations for the Supervised Fine-Tuning (SFT) and Reinforcement Learning (RL) stages are summarized in Table~\ref{tab:sft_config} and Table~\ref{tab:rl_config}, respectively.

To ensure a fair and rigorous comparison, we standardize the input configurations across all evaluated models during the inference stage, including both our proposed method and other open-source baselines. 
Specifically, we increase the temporal density by maintaining the video length at 32 frames. 
Correspondingly, the frame resolution is set to $256 \times 28 \times 28$ pixels to capture finer visual details for precise progress estimation.

\begin{table}[t]

\centering
\caption{SFT Training Config}
\label{tab:sft_config}
\begin{tabular}{p{5cm} p{7cm}}
\hline
\textbf{Configuration} & \textbf{Value} \\


\hline
\multicolumn{2}{l}{\textbf{Algorithm}} \\
trainer & TRL SFTTrainer \\
seed & 42 \\

\hline
\multicolumn{2}{l}{\textbf{Model}} \\
freeze\_vision\_tower & FALSE \\
enable\_gradient\_checkpointing & TRUE \\
attn\_implementation & flash\_attention\_2 \\
precision & bf16 \\

\hline
\multicolumn{2}{l}{\textbf{Batching}} \\
nproc\_per\_node & 8 \\
per\_device\_train\_batch\_size & 1 \\
gradient\_accumulation\_steps & 8 \\
global\_batch\_size & $1 \times 8 \times 8 = 64$ \\

\hline
\multicolumn{2}{l}{\textbf{Optimization}} \\
strategy & adamw \\
lr  & $1.0e^{-6}$  \\
weight\_decay & 0.0 \\
lr\_warmup\_ratio  & 0.0 \\

lr\_scheduler\_type  & linear \\
num\_train\_epochs & 1 \\

\hline
\multicolumn{2}{l}{\textbf{Rollout / inference}} \\
num\_generations	& 8 \\
max\_turns & 3 \\
top\_p / temperature & 0.9 / 0.7 \\

\hline

\end{tabular}
\end{table}

\begin{table}[t]
\centering
\caption{RL Training Config}
\label{tab:rl_config}
\begin{tabular}{p{5cm} p{7cm}}
\hline
\textbf{Configuration} & \textbf{Value} \\
\hline

\multicolumn{2}{l}{\textbf{Data constraints}} \\
min\_pixels / max\_pixels & 3136 / 401408 \\
max\_prompt\_length  &  16384  \\
max\_completion\_length &  4096 \\

\hline
\multicolumn{2}{l}{\textbf{Algorithm}} \\
algorithm & grpo \\
reward functions & accuracy\_reward + format\_reward \\
kl\_coef (beta) & 0.04 \\
temporal / len\_control & false / true \\

\hline
\multicolumn{2}{l}{\textbf{Model}} \\
freeze\_vision\_tower & FALSE \\
enable\_gradient\_checkpointing & TRUE \\
attn\_implementation & flash\_attention\_2 \\
precision & bf16 \\

\hline
\multicolumn{2}{l}{\textbf{Batching}} \\
per\_device\_train\_batch\_size & 1 \\
gradient\_accumulation\_steps & 1 \\
number of GPUs & 4 \\
global prompt batch & 4 \\
num\_generations (G) & 8 \\
rollouts per step & 32 \\
max\_grad\_norm & 5 \\

\hline
\multicolumn{2}{l}{\textbf{Optimization}} \\
strategy & adamw \\
lr   & $1.0e^{-6}$  \\
weight\_decay & 0.01 \\
lr\_warmup\_ratio & 0.0 \\

lr\_scheduler\_type & cosine  \\
num\_train\_epochs & 1 \\

\hline
\multicolumn{2}{l}{\textbf{Rollout / Inference}} \\
n (generations views) & 8 \\
generation: do\_sample & TRUE \\
generation: max\_new\_tokens & 4096 \\
generation: top\_p / temperature & 0.95 / 1.0 \\

\hline
\end{tabular}
\end{table}

\end{document}